\documentclass[runningheads]{llncs}
\usepackage[final,year=2026,ID=13080]{eccv}
\usepackage{eccvabbrv}
\usepackage[breaklinks,colorlinks,citecolor=eccvblue]{hyperref}
\usepackage{orcidlink}

\usepackage{graphicx}
\usepackage{booktabs}
\usepackage{amssymb}
\usepackage{pifont}
\usepackage{amsmath}
\usepackage{bm}
\usepackage{comment}
\usepackage{balance}
\usepackage{tabularx}
\usepackage{xcolor}
\usepackage{enumitem}
\usepackage{tikz}
\usepackage{pgf-pie}
\usepackage{caption}
\usepackage{subcaption}
\usepackage{multirow}
\usepackage[most]{tcolorbox}
\usepackage{siunitx}
\usepackage{float}
\usepackage{makecell}
\usepackage[table]{xcolor}
\usepackage{marvosym}

\providecommand{\model}[1]{\texttt{#1}}

\definecolor{banglaGreen}{RGB}{0,120,70}     % Bangla better
\definecolor{englishBlue}{RGB}{0,70,140}     % English better
\definecolor{neutralPink}{RGB}{220,90,140} 

\newcommand{\ban}{{\large$\bullet$}}
\newcommand{\eng}{\footnotesize$\blacksquare$}
\newcommand{\eqv}{\footnotesize$\blacklozenge$}

\newcommand{\cmark}{\ensuremath{\checkmark}}

\lstdefinestyle{prompt}{
    basicstyle=\footnotesize,
    breaklines=true,
    breakatwhitespace=true,
    breakindent=0pt,
    breakautoindent=false,
    columns=fullflexible,
    keepspaces=true,
    showstringspaces=false,
    mathescape=true,
    moredelim=**[is][\bfseries]{|}{|}
  }
  \definecolor{titlebar}{HTML}{6B7A3A}
  \newtcblisting{promptbox}[1]{
    colback=white, colframe=titlebar, coltitle=white, colbacktitle=titlebar,
    fonttitle=\bfseries\small, title={#1},
    arc=2pt, boxrule=0.6pt,
    left=6pt, right=6pt, top=4pt, bottom=4pt,
    enhanced, breakable, listing only,
    listing options={style=prompt},
  }

\begin{document}

    \title{\textbf{BaFCo}: A Document Understanding Benchmark for Complex \underline{\textbf{Ba}}ngla \underline{\textbf{F}}orm \underline{\textbf{Co}}mprehension}
    
    \titlerunning{BaFCo Benchmark}
    \authorrunning{A. Azad et al.}

    \author{
        Abu Tyeb Azad\inst{1,\star}\textsuperscript{(\Letter)} \and
        Ishita Sur Apan\inst{2,\star} \and
        Fahim Ahmed\inst{2,\star} \and
        Sumaiya Karim Katha\inst{2} \and
        Ezharuddin Jubaer\inst{2} \and
        Armun Alam\inst{2} \and
        Pranjal Kumar Nandi\inst{3} \and
        Amin Ahsan Ali\inst{2} \and
        Aman Chadha\inst{4,\ddagger} \and
        Md Mofijul Islam\inst{4,\dagger,\ddagger} \and
        AKM Mahbubur Rahman\inst{2,\dagger}
    }
    \institute{
        Wichita State University, USA
        \and
        Center for Computational \& Data Sciences, Bangladesh
        \and
        University of Dhaka, Bangladesh
        \and
        Amazon GenAI, USA 
    }

    \maketitle
    {\let\thefootnote\relax
    \footnotetext{
        $^{\star}$Equal contribution.\quad
        $^{\dagger}$Equal supervision.\quad
        $^{\ddagger}$Work done outside role at Amazon.\\
        {(\Letter)}~Corresponding author: \email{mausulazad495@gmail.com}}
    }
    {\let\thefootnote\relax
    \footnotetext{
        \textit{\\Accepted at the 19th European Conference on Computer Vision (ECCV), 2026.}
    }
    
    \begin{abstract}
        Document comprehension is a challenging yet impactful task for Multimodal Large Language Models, especially as these systems see growing adoption in real-world, human-centric applications. However, this adoption is limited for low-resource languages such as Bangla due to the scarcity of high-quality annotated data. To address this gap, we introduce \textbf{BaFCo}, a benchmark dataset for Bangla form comprehension with a focus on Document Layout Analysis (DLA) and Key Information Extraction (KIE). BaFCo curates 200 multi-page complex Bangladeshi government forms, sourced from across diverse sectors including agriculture, education, banking, and land management. To accurately capture the structural and contextual complexity of these forms, we define a fine-grained annotation schema comprising 26 types of form entities, along with a separate coarse form entity set consisting of 5 types. We evaluate the latest MLLMs from the ChatGPT, Gemini, Claude, Qwen, and Kimi series using zero-shot and chain-of-thought prompts under both low and high reasoning setups.  Our results reveal limitations in current MLLMs' ability in comprehending Bangla forms, particularly in accurately localizing highly granular form entities. Our dataset and code is available at: \url{https://huggingface.co/datasets/Mausul/bafco} 
        \keywords{Document Comprehension, Document Layout Analysis, Key Information Extraction, Low-resource NLP, Multimodal LLM}
    \end{abstract}
    
    \section{Introduction}
    Information retrieval from documents such as forms underpins many real-world systems including banking, education, and public administration \cite{un_report}. \textbf{Document Layout Analysis} (DLA) and \textbf{Key Information Extraction} (KIE) \cite{dla, kie_survey_1, kie_survey_2} are tasks at the core of Document Understanding. DLA identifies the structural elements of a page and the relationships between them. KIE involves locating and extracting values from form fields filled digitally or by hand. Both DLA and KIE are foundational precursors to downstream document tasks such as document question answering, entity linking, and summarization.

Bangla remains a low-resource language in document understanding despite being the world's 7th most spoken language, with 284 million speakers \cite{zaban}. Government forms are also underrepresented in research, despite their semantic diversity and practical importance in public services. Benchmarks such as FUNSD \cite{funsd} and XFUND \cite{xfund} have advanced form understanding in English and other languages, but no comparable benchmark has been available for Bangla forms. Consequently, the lack of high-quality datasets and benchmarks continues to limit the development of Bangla document understanding systems.

To address this gap, we introduce \textbf{BaFCo}, a curated benchmark for multi-page Bangla form comprehension, focusing on government forms and the tasks of DLA and KIE. BaFCo provides fine-grained annotations spanning 26 entity types and labeled relationships between related fields. For DLA, the dataset contains 16,382 entities and 8,771 relationships across 200 forms (316 pages, with 1--5 pages per form). For KIE, it further includes 1,926 key-value pairs spanning 156 forms (186 pages). To maximize diversity, we prioritize complex and varied layouts over simpler forms. All annotations are created by trained annotators and reviewed by experts to ensure structural and semantic quality. In addition, we provide an end-to-end evaluation pipeline, standard document-comprehension metrics, and a coarse label set with five entity types, enabling systematic evaluation of MLLMs on Bangla DLA and KIE tasks.

With the advent of LLMs and MLLMs, document understanding workflows have increasingly shifted toward generative approaches, driven in part by the popularity of conversational interfaces. MLLMs are now widely used for document understanding tasks \cite{genkie, genkie2, genkie3}, replacing earlier OCR-based methods \cite{ofa}. Although document-specialized models are available \cite{docllm, layoutllm, mplug, dockylin}, off-the-shelf flagship MLLMs offer a compelling alternative due to their lower development overhead through publicly accessible APIs, the scarcity of high-quality domain-specific training data, and the substantial computational resources required to train custom models \cite{llm_cost, mllm_cost}.

Consequently, we evaluate flagship MLLMs on BaFCo using tuning-free prompt-based methods to assess their layout-grounding capabilities and their understanding of Bangla-specific form elements. In DLA, \model{Gemini 3 Pro} performs the best across all experimental setup with average mAP scores of 0.1177 and 0.2646 for granular and coarse entity set, respectively. For KIE, \model{Gemini 3 Pro} again outperforms other models for Bangla forms, but for English ones \model{GPT-5.2} is surpassing others. We also evaluate the models on a set of English forms from the same domains to examine the effects of language. In case of DLA (particularly for granular entity set), the effect is minimal (performance difference $\le 0.02$), whereas for KIE \model{GPT-5.2} and \model{Claude Opus 4.6} performs better on English forms and \model{Gemini 3 Pro} on Bangla forms. 

\begin{figure*}[!bht]
    \centering
    \includegraphics[width=\textwidth]{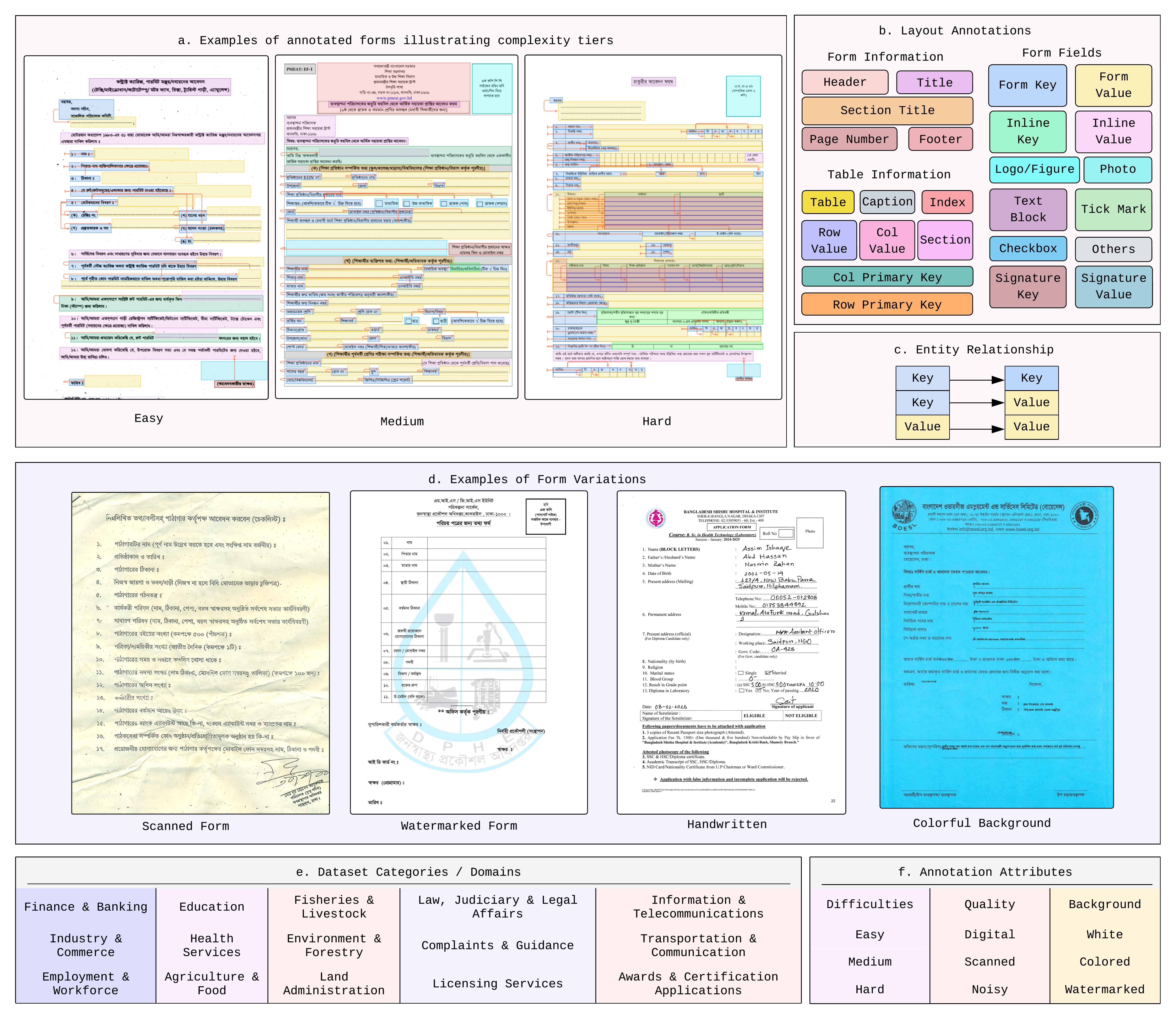}
    
    \caption{Overview of BaFCo data diversity. (a) Examples of annotated forms across three difficulty levels: easy, medium, and hard (see \cref{dataset_desc}); (b) BaFCo supports detailed layout annotations with 26 entity types, including titles, form key-value pairs, and tables; (c) It contains links between related entities through three relationship types: key-to-key, key-to-value, and value-to-value; (d) Examples of form variations, including scanned, watermarked, and handwritten forms; (e) BaFCo includes government forms from 15 domains; (f) Each form is annotated with nine page-level attributes based on annotation difficulty, image quality, and form background.}
    \label{fig:bafco}
\end{figure*}

Our main contributions are as follows:
\begin{enumerate}
    \item We introduce BaFCo, a high-quality dataset of Bangladeshi government forms with 26 entity types and annotated relationships between related fields, validated through human annotation and expert review.
    
    \item We present the first benchmark for Document Layout Analysis (DLA) and Key Information Extraction (KIE) on Bangla forms, addressing a critical gap in low-resource document understanding.
    
    \item We provide an empirical analysis of the performance of off-the-shelf flagship MLLMs on DLA and KIE tasks for Bangla documents, revealing limitations of the current models, particularly in handling fine-grained layouts.
\end{enumerate}

    \section{Related Works}
    \textbf{Document Layout Analysis (DLA)} aims to identify and localize structural elements, such as text blocks, tables, images, and form fields. Early benchmarks spurred research from rule-based systems \cite{ha1995recursive, journet2005text} to machine learning \cite{wu2008machine, bukhari2010document} for segmenting and classifying document regions. However, they struggle with complex real-world layouts. Hence, recent approaches predominantly rely on deep learning models \cite{docopilot, doclayllm, layoutlmv3, mplug, publaynet} that combine visual, textual, and structural cues, including CNN-Transformer hybrids and multimodal architectures, which require high-quality annotations to generalize across varied domains and document types. Early foundational datasets like PubLayNet \cite{publaynet} and DocBank \cite{li2020docbank} scaled through leveraging scientific articles from PubMed Central and arXiv, respectively. However, these born-digital scientific datasets lack real-world layout variability. DocLayNet \cite{pfitzmann2022doclaynet} addressed this gap by providing data from diverse sources; including financial reports, manuals, and legal documents.

\textbf{Key Information Extraction (KIE)} focuses on identifying and extracting form key-value pairs. Among generative KIE approaches,  GenKIE \cite{genkie} uses an encoder-decoder architecture with zero-shot prompt based evaluation. In BROS \cite{bros} uses a multimodal, layout-aware encoder. LiLT \cite{lilt} and OmniDocBench \cite{omnidocbench} are multilingual, nonetheless Bangla is absent from their pool. Apart from GPT4o in OmniDocBench, no work evaluates recent flagship MLLMs' on KIE.   

\textbf{Form Understanding} is an important subfield of document intelligence, focusing on extracting, structuring, and linking information from forms. FUNSD \cite{funsd} was among the first public benchmarks in this area, providing annotations for questions, answers, headers, and other text entities, along with question--answer relationships. It enabled research on semantic entity recognition and relation extraction, but its annotation schema is largely limited to flat question--answer structures. Subsequent datasets have targeted specific domains and document types, including government business documents \cite{buddie}, financial forms \cite{formnlu}, receipts \cite{sroie, cord, docile}, and legal and financial documents \cite{kleister}. While these datasets contain challenging layouts and long-form content, they generally lack fine-grained text-entity annotations and spatial relationships required for detailed form structure understanding, particularly in low-resource and domain-specific settings.

\textbf{Multilingual and Low-Resource Document Datasets}
Most document layout analysis benchmarks focus on high-resource languages, leaving low-resource settings underexplored. XFUND \cite{xfund} introduced a multilingual benchmark covering seven languages. However, many low-resource languages remain absent. Bangla is one such example: despite its large speaker base, public benchmarks for Bangla document understanding have been scarce. A first step was BaDLAD \cite{badlad}, which introduced a multi-domain dataset spanning books, newspapers, government documents, and historical records. Although it provides 710k polygon annotations, its labels are limited to coarse layout elements such as text regions, paragraphs, tables, and images. However, it does not support fine-grained form understanding with form-specific entities, key--value relationships, or government-form structures. 

\begin{table}[tbh]
    \centering
    % \caption{Comparison between \textbf{BaFCo} and existing document understanding benchmarks. \textbf{Domain}: Number of document domains (e.g. Finance, Education, Health, etc) covered in the dataset. \textbf{Form}: Presence of form-style documents. \textbf{LRL}: Low-resource language support. \textbf{Multipage}: Multi-page document annotations. \textbf{Human}: Human-annotated dataset. \textbf{BBox}: Bounding box annotations of the labels. \textbf{Text}: Text field annotations. Tab: Table annotations. \textbf{Img}: Images such as figures, logos, or diagrams. \textbf{Sig}: Signature annotations. \textbf{Cb}: Checkbox annotations including single-select, multi-select, and radio buttons. \textbf{Photo}: Attached photographs of the form holder. \textbf{EL}: Entity linking or relationships between document fields. \textbf{Tasks}: Document Layout Analysis and Key Information Extraction.}
    
    \caption{Comparison of \textbf{BaFCo} with existing document understanding benchmarks. \textbf{Domain} denotes the no. of document domains (e.g., finance, education, healthcare). \textbf{Form} indicates support for form-style documents; \textbf{LRL}, low-resource languages; \textbf{Multipage}, multi-page annotations; and \textbf{Human}, human-created annotations. \textbf{BBox} and \textbf{Text} denote bounding-box and text annotations, respectively. \textbf{Tab}, \textbf{Img}, \textbf{Sig}, \textbf{Cb}, and \textbf{Photo} indicate annotations for tables, images (e.g., figures, logos, diagrams), signatures, checkboxes (including single-select, multi-select, and radio buttons), and photographs. \textbf{EL} denotes entity linking between related fields. \textbf{Tasks} lists supported tasks: Document Layout Analysis (DLA) and Key Information Extraction (KIE).}

    \label{tab:benchmark-summary}

    \setlength{\tabcolsep}{3.2pt}
    \renewcommand{\arraystretch}{1.18}

    \resizebox{0.9\textwidth}{!}{%
        \begin{tabular}{l|ccc|cccccccccc|cc}
            \toprule
            \multirow{2}{*}{\textbf{Benchmark}} &
            \multicolumn{3}{c|}{\textbf{Data}} &
            \multicolumn{10}{c|}{\textbf{Annotation Attributes and Types}} &
            % \multicolumn{5}{c|}{\textbf{Annotation Type}} &
            \multicolumn{2}{c}{\textbf{Tasks}} \\
            % \cline{2-3}\cline{4-6}\cline{7-11}\cline{12-13}\cline{14-16}
            & {Domain} & {Form} & {LRL}
            & {Multipage} & {Human}
            & {BBox} & {Text} & {Tab} & {Img} & {Sig} 
            & {Cb} & {Photo} & {EL}
            & {DLA} & {KIE} \\
            \midrule
            
            \multicolumn{16}{l}{\textit{DLA Benchmarks}}\\
            \midrule
            
            \makecell[l]{PubLayNet~\cite{publaynet}} &
            2 & -- & -- &  
            -- & -- &
            \cmark & \cmark & \cmark & \cmark & 
            -- & -- & -- & -- & 
            \cmark & -- \\
            % \midrule
            
            \makecell[l]{DocBank~\cite{li2020docbank}} &
            1 & -- & -- &  
            \cmark & -- &
            \cmark & \cmark & \cmark & \cmark & 
            -- & -- & -- & -- & 
            \cmark & -- \\
            % \midrule
            
            \makecell[l]{DocLayNet~\cite{pfitzmann2022doclaynet}} &
            5 & -- & -- &  
            -- & \cmark &
            \cmark & \cmark & \cmark & \cmark & 
            -- & -- & -- & -- & 
            \cmark & -- \\
            % \midrule
            
            \makecell[l]{OmniDocBench~\cite{omnidocbench}} &
            9 & -- & -- &  
            \cmark & \cmark &
            \cmark & \cmark & \cmark & \cmark & 
            -- & -- & -- & -- & 
            \cmark & -- \\
            \midrule
            
            \multicolumn{16}{l}{\textit{Form Benchmarks}}\\
            \midrule

            BuDDIE~\cite{buddie} &
            3 & \cmark & -- &  
            -- & \cmark &
            \cmark & \cmark & -- & -- & 
            -- & -- & -- & -- & 
            -- & \cmark \\

            FUNSD~\cite{funsd} &
            1 & \cmark & -- &  
            -- & \cmark &
            \cmark & \cmark & -- & -- & 
            -- & -- & -- & \cmark & 
            \cmark & \cmark \\

            XFUND~\cite{xfund} &
            1 & \cmark & -- & 
            -- & \cmark &
            \cmark & \cmark & -- & -- & 
            -- & -- & -- &
            \cmark & 
            -- & \cmark \\
            % \midrule
            
            FormNLU~\cite{formnlu} &
            1 & \cmark & -- & 
            -- & \cmark &
            \cmark & \cmark & -- & -- & 
            -- & -- & -- &
            \cmark & 
            \cmark & \cmark \\
            % \midrule
            
            \midrule
            
            \multicolumn{16}{l}{\textit{Low-Resource Language Benchmark}}\\
            \midrule
            
            BaDLAD~\cite{badlad} &
            4 & -- & \cmark & 
            -- & \cmark &
            \cmark & \cmark & \cmark & \cmark & 
            -- & -- & -- &
            -- & 
            \cmark & -- \\
            % \midrule
            \midrule
            
            \rowcolor{black!10}
            \textbf{BaFCo (ours)} &
            15 & \cmark &
            \cmark & \cmark & \cmark &
            \cmark & \cmark & \cmark & \cmark & \cmark &
            \cmark & \cmark &
            \cmark & \cmark & \cmark \\
            \bottomrule
        \end{tabular}%
    }
\end{table}

% \makecell[l]{
%     BuDDIE~\cite{buddie}\\
%     FUNSD~\cite{funsd}\\
%     XFUND~\cite{xfund}\\
%     FormNLU~\cite{formnlu}
% } & 
%  &  &
% \makecell{
%     \cmark\\ \\ \\ \cmark
% } & 
% \makecell{
%     \cmark\\\cmark\\\cmark\\ \cmark
% } & 
% \makecell{
%     \\\cmark\\\cmark\\ \cmark
% } & 
% \makecell{
%     \cmark\\ \\ \\ \cmark
% } & 
% \makecell{
%     \\ \\ \\ \cmark
% } & 
% \makecell{
%     \\ \\ \\ \cmark
% } & 
% \makecell{
%     \\ \\ \\ 
% } & 
% \makecell{
%     \\ \\ \\ 
% } & 
% \makecell{
%     \\ \\ \\ 
% } & 
% \makecell{
%     \cmark\\\cmark\\\cmark\\ \cmark
% } & 
% \makecell{
%     \\\cmark\\\cmark\\ \cmark
% } & 
% \makecell{
%     \\\cmark\\\cmark\\ 
% } \\

Our work addresses this gap with BaFCo, the first publicly available benchmark for Bangla form comprehension and layout analysis. BaFCo provides fine-grained annotations spanning 26 entity types and labeled relationships between related entities (e.g., linking field labels to values), a capability not previously available for Bangla documents.
    
    \section{Dataset}
    \subsection{Data Collection and Curation}

We adopt a quality-over-quantity strategy, prioritizing careful selection, diversity, and annotation fidelity over scale. The dataset is composed of publicly available Bangladeshi government forms\footnote{\url{https://forms.portal.gov.bd/}}, ensuring authenticity and real-world relevance. Government forms were chosen due to their structured yet highly variable layouts, dense semantic content, and practical importance.

Forms were filtered to remove duplicates, incomplete scans, and low-quality images that could introduce annotation noise. The selected forms span multiple administrative domains, including taxation, healthcare, education, and civil services. They exhibit varying degrees of layout complexity, including multi-column structures, nested fields, tables, and handwritten or stamped regions. We further categorize forms into three difficulty levels-easy, medium, and difficult-based on layout density, visual clutter, and semantic coupling between fields.

Unlike large-scale multilingual benchmarks such as OmniDocBench \cite{omnidocbench} or BuDDIE \cite{buddie}, which emphasize breadth across document types and languages, our dataset is purpose-built for Bangla forms. Accordingly, comparisons in this work focus on Bangla-specific and multilingual datasets that include Bangla content. The dataset is intended as a high-quality benchmark for low-resource document layout analysis rather than a general-purpose corpus.

\subsection{Dataset Description}
\textbf{Form Selection Criteria and Difficulty Levels:} \label{dataset_desc}
Forms containing 1--5 pages were first shortlisted. The selected forms were then grouped into difficulty levels based on layout complexity and component diversity. \textit{Easy} forms contained basic elements such as form key--value pairs (including one-to-many and key-to-key relationships), inline and signature key--value pairs, headers, titles, section headings, text blocks, photo fields, and footers. However, they did not include tables, checkboxes, or tick marks. \textit{Medium} forms introduced additional structural complexity through checkboxes, tick marks, and tables consisting only of columns, while explicitly excluding tables with both rows and columns, sub-columns, or embedded checkboxes and tick marks. \textit{Hard} forms represented the highest level of complexity and included tables with rows, columns, and sub-columns; tables containing checkboxes or tick marks; and other unique, dense, or unconventional layout components.

\textbf{DLA Dataset Composition:}
Following the difficulty-based grouping, BaFCo encompasses a diverse range of form structures, including application and non-application forms, sparse and dense layouts (ranging from forms with few fields to those containing crowded or heavily tabular regions), and both guided forms (with explicit bounding boxes) and unguided forms (without such cues). These variations are representative of real-world government forms. BaFCo also exhibits a deliberately skewed entity-class distribution, preserving the natural long-tail distribution of entities found in real-world documents rather than imposing artificial class balance. Common elements (e.g., form keys and values) dominate the dataset, whereas specialized elements (e.g., signatures and tabular sub-fields) are relatively rare, reflecting the inherent long-tailed nature of government forms. Overall, the dataset contains 16,382 entities and 8,771 relationships across 200 forms (316 pages), with each form comprising between 1 and 5 pages.

\textbf{KIE Dataset Composition:}
For KIE, we evaluate 156 forms (186 pages and 1,926 key--value pairs) spanning the same Easy, Medium, and Hard difficulty levels. English forms are included to facilitate comparisons between a high-resource language and a low-resource language. The Bangla-to-English ratio is 1.14:1 by form count and 1.24:1 by key--value pair count.

% Overall STAT
\begin{table}[htb]
    \centering
    \caption{BaFCo dataset statistics for the DLA and KIE tasks, broken down by difficulty level. Under our difficulty rubric, easy forms do not contain tables or selection elements (checkboxes/tick marks); consequently, these fields are absent from the Easy category and are denoted by `--'.}
    
    \label{tab:annotation_stats}
    \begin{minipage}[t]{0.5\linewidth}
        \centering
        \resizebox{\textwidth}{!}{ 
            \begin{tabular}{lrrrr}
                \toprule
                \textbf{} & \textbf{Easy} & \textbf{Medium} & \textbf{Hard} & \textbf{Total} \\ 
                \midrule
                % \multicolumn{5}{l}{\textbf{\textit{Dataset Size}}} \\
                % Forms                & 74 & 66 & 60 & 200 \\
                % Pages                & 90 & 108 & 118 & 316 \\
                % Tables               & 0 & 59 & 143 & 202 \\
                % \\
                % \multicolumn{5}{l}{\textbf{\textit{Structure}}} \\
                % Avg. entities/form    & 55.76 & 77.76 & 118.73 & - \\
                % {Avg. relations/form   }      & 30.01 & 37.74 & 67.65 & - \\
                % \midrule
                % \\

                \multicolumn{5}{l}{\textbf{\textit{Dataset Size}}} \\
                \quad DLA Forms & 74 & 66 & 60 & 200 \\
                \quad DLA Pages & 90 & 108 & 118 & 316 \\
                \quad DLA Tables & -- & 59 & 143 & 202 \\
                \quad KIE Forms & 86 & 42 & 28 & 156 \\
                \quad KIE Pages & 95 & 57 & 34 & 186 \\
                
                \multicolumn{5}{l}{\textbf{\textit{Structure}}} \\
                \quad DLA Avg. entities/form & 55.76 & 77.76 & 118.73 & 81.91 \\
                \quad DLA Avg. relations/form & 30.01 & 37.74 & 67.65 & 43.85 \\
                \quad KIE Avg. key-value pairs/form & 11.2 & 13.5 & 15.0 & 12.3 \\
                \bottomrule
            \end{tabular}
        }
    \end{minipage}%
    \begin{minipage}[t]{0.5\linewidth}
        \centering
        \resizebox{.92\textwidth}{!}{ 
            \begin{tabular}{lrrrr}
                \toprule
                \textbf{} & \textbf{Easy} & \textbf{Medium} & \textbf{Hard} & \textbf{Total} \\ 
                \midrule
                % \multicolumn{5}{l}{\textbf{\textit{Annotation Types}}} \\
                % Form values          & 1,418 & 1,179 & 1,079 & 3,676 \\
                % Inline values        & 155 & 198 & 216 & 569 \\
                % Signatures           & 187 & 189 & 144 & 520 \\
                % Checkboxes           & 0 & 245 & 166 & 411 \\
                % Tick marks           & 0 & 65 & 51 & 116 \\
                % Table rows           & 0 & 313 & 622 & 935 \\
                % Table columns        & 0 & 262 & 963 & 1,225 \\
                % \midrule
                % \textbf{Total annotations  } & \textbf{1,760} & \textbf{2,451} & \textbf{3,241} & \textbf{7,452}\\

                \multicolumn{5}{l}{\textbf{\textit{Annotation Types}}} \\
                \quad DLA Form values & 1,418 & 1,179 & 1,079 & 3,676 \\
                \quad DLA Inline values & 155 & 198 & 216 & 569 \\
                \quad DLA Signatures & 187 & 189 & 144 & 520 \\
                \quad DLA Checkboxes & -- & 245 & 166 & 411 \\
                \quad DLA Tick marks & -- & 65 & 51 & 116 \\
                \quad DLA Table rows & -- & 313 & 622 & 935 \\
                \quad DLA Table columns & -- & 262 & 963 & 1,225 \\
                \quad DLA Total annotations & 1,760 & 2,451 & 3,241 & 7,452 \\
                \quad KIE Filled key-value pairs & 955 & 568 & 391 & 1,926 \\
                \bottomrule
            \end{tabular}
        }
    \end{minipage}
\end{table}

\subsection{Semantic Form Entity Definition}

Based on our analysis of Bangladeshi government forms, we defined a fine-grained and comprehensive taxonomy of semantic form entities. The dataset contains 26 entity types covering structural components (e.g., headers, sections, and tables), functional fields (e.g., key--value pairs, checkboxes, and signatures), and auxiliary elements (e.g., instructions, seals, and stamps) across both single-page and multi-page forms (see \cref{tab:layout_categories}). To examine the effect of entity granularity on model performance, we additionally group these 26 entity types into five coarse-grained categories.

In addition to entity labels, we annotate relationships between semantically linked fields, enabling the use of BaFCo for downstream document understanding tasks. \cref{tab:annotation_stats} summarizes the total number of annotated entities and relationships across the different difficulty levels. The taxonomy for each entity type was iteratively refined through pilot annotations and expert feedback to ensure coverage, consistency, and applicability across diverse form layouts. This detailed taxonomy distinguishes our dataset from existing Bangla document resources, which typically employ generic or task-specific entity label sets.

\begin{table}[htb]
    \centering
    \caption{Semantic Form Entities. Both coarse and granular form entities are enlisted.}
    \label{tab:layout_categories}
    \resizebox{0.75\textwidth}{!}{
        \begin{tabular}{@{}p{0.15\linewidth} p{0.75\linewidth}@{}}
            \toprule
            \textbf{Coarse} & \textbf{Granular} \\ 
            \midrule
            
            Headers & Header, Title, Footer, Section Title \\  
            \midrule
            
            Table & Table, Table Caption, Table Section, Table Index, Row Primary Key, Row Value, Column Primary Key, Column Value \\  
            \midrule
            
            Image & Diagram / Logo / Figure \\  
            \midrule
            
            Fields & Form Key, Form Value, Inline Key, Inline Value, Signature Key, Signature Value, Tick Mark, Checkbox, Image Field \\  
            \midrule
            
            Others & Text Block, Page Number, Gibberish, Others \\ 
            \bottomrule
        \end{tabular}
    }
\end{table}

\subsection{Annotation Guidelines}

\subsubsection{DLA:}
To ensure high annotation quality and reproducibility, we developed a detailed annotation guideline document that specifies entity definitions, bounding box rules, relationship constraints, and edge cases. Particular emphasis is placed on resolving ambiguities common in Bangla forms, such as overlapping text regions, visually implicit field boundaries, and mixed printed-handwritten content. The guideline is refined through multiple rounds of annotator feedback and validation, resulting in a consistent annotation protocol that minimizes subjective interpretation while retaining flexibility for complex layouts.

\subsubsection{KIE:}
For KIE, we annotate bounding boxes and define explicit relationships between regions. Two labels --- \textit{key} and \textit{value} --- were used. Annotators filled a text field for each region: key fields copied the annotated text, while value fields recorded the corresponding entry. These fields served as ground truth. To evaluate models, values were overlaid on form images, and LLMs were prompted to extract them given the associated keys. To assess model performance, the values were superimposed onto the form images, and the LLMs were prompted to extract the values given the corresponding keys.

\subsection{Annotation Procedure}
17 annotators underwent two days of training using guideline documents, practice forms, and tutorial videos covering dataset nuances. All were proficient in Bangla and experienced in document labeling. Each document was independently annotated in \textit{Label Studio}\footnote{\url{https://labelstud.io/}} by a trained annotator and reviewed by an expert reviewer. Disagreements or uncertainties in labels or bounding boxes were resolved via group discussion and majority voting. To quantify reliability, we measured agreement between the trained annotator and the expert reviewer on the pre-majority-vote annotations, obtaining a Cohen's $\kappa$ of $0.974$.

\subsection{Difficulties Faced During Annotation:} 
Most annotation errors stemmed from layout complexity and fell into three categories. \textit{Semantic confusions} arose when distinguishing closely related entity types: in fill-in-the-blank forms, standard key--value pairs were frequently mislabeled as inline or signature pairs, and slashes (``/'') were occasionally mistaken as separators between Tick Mark entities. \textit{Structural ambiguities} were most common in guided layouts, where form keys were hard to separate from table cells, and where a few multi-cell tables were annotated cell-by-cell rather than as a single structure. \textit{Technical inconsistencies} included misclassification, duplicate boxes, and overly broad inline annotations, along with occasional omissions of essential elements (e.g., header logos, signature links) and erroneous inclusion of irrelevant footer text. All these issues were identified and corrected by expert reviewers prior to finalization.
    
    \section{Experiments}
    \subsection{Experimental Setup}

\subsubsection{Models:}
\label{subsec:models}
We evaluate five flagship MLLMs spanning both proprietary and open-source ecosystems: \model{GPT-5.2}, \model{Gemini 3 Pro}, and \model{Claude Opus 4.6} on the proprietary side, and \model{Qwen 3.6 Plus} and \model{Kimi K2.5} among open-source models. We focus exclusively on MLLMs because they represent the most promising paradigm for general-purpose document understanding. Unlike OCR-based pipelines and encoder architectures, which are typically designed for predefined tasks with fixed output schemas, MLLMs operate in an open-ended, instruction-driven manner that better aligns with emerging document AI applications. Consequently, direct comparisons with OCR or encoder-based systems are less informative for our objectives. We therefore center our evaluation on MLLMs, which are also widely accessible through API-based deployment. For inference, we use native batch endpoints for proprietary models to improve cost efficiency and \textit{OpenRouter}\footnote{\url{https://openrouter.ai/}} endpoints for open-source models.

\subsubsection{Prompts:}
To utilize MLLMs' inference-time task-completion capabilities, we used zero-shot \cite{zero_shot} and Chain-of-Thought (CoT) prompting \cite{cot}. We included task descriptions, form entity specifications, and output-structure instructions in the prompts. For KIE, we followed the prompt structure of GenKIE \cite{genkie}, where KIE is formulated as a visual question-answering task.

\subsubsection{Reasoning Effort:}
Besides CoT prompting, for DLA experiments we also use the built-in reasoning of the flagship models. We run every setup at two reasoning levels, \emph{low} and \emph{high}, set via the \emph{reasoning\_effort} API parameter. We use this as our main control because it is the only setting shared across providers for adjusting models' reasoning level. Since black-box APIs do not reveal each model's internals, it is the fairest basis for comparison. To cap runaway generation, we limit \emph{max\_output\_tokens} at $16{,}000$ for low reasoning and $64{,}000$ for high reasoning. For consistency, we use only the \emph{low} and \emph{high} levels, as these settings were available across all providers (\model{Gemini 3 Pro}'s API did not offer a \emph{medium} reasoning effort level at the time of writing).

\subsubsection{DLA Evaluation Metrics:}
\label{subsec: metrics}
We evaluate model predictions using standard detection and classification metrics following prior form understanding and object detection benchmarks \cite{formnlu, omnidocbench, coco}. As in \cite{formnlu, omnidocbench}, we perform greedy matching between predicted and ground-truth bounding boxes with two Intersection-over-Union (IoU) thresholds $\tau \in {0.3, 0.5}$ to ensure order-invariant assignment.

A predicted box is counted as a true positive (TP) if matched to a ground-truth instance with $\mathrm{IoU} \ge \tau$. Unmatched predictions are false positives (FP), and ground-truth instances without corresponding predictions are false negatives (FN). Using these counts, we compute precision, recall, F1 score, and mean average precision (mAP). For a fixed IoU threshold $\tau$, mAP is obtained by averaging class-wise Average Precision (AP) across all form field categories.

\subsubsection{KIE Evaluation Metrics:}
Following prior work \cite{formnlu, omnidocbench, genkie}, we use precision, recall, F1, and Normalized Edit Similarity ($NES$) for KIE evaluation. 

$NES$ is defined as $NES = 1 - NED$, where $NED = \frac{d(s,t)}{\max(|s|,|t|)}$ and $d(s,t)$ is the Levenshtein distance between the predicted string $s$ and ground-truth $t$.

\subsection{Evaluation Setup}
To evaluate MLLMs' performance for each task $t \in \mathit{T}$ where $\mathit{T} = \left\{ DLA,\; KIE \right\}$, image of page $j$ of form $i$, $\mathbf{b}_{i,j}$ is selected from BaFCo dataset $\mathit{B}$. Then along with prompt $p$ ($p \in \mathit{P_t}$) the page image $\mathbf{b}_{i,j}$ is passed to each MLLM, $M_{\theta}$ from a pool of models (see \cref{subsec:models}),to generate raw prediction $\tilde{y}_{i,j}^{\,p}$.

\begin{equation}
\tilde{y}_{i,j}^{\,p} = M_{\theta}\!\left(\mathbf{b}_{i,j},\, p\right)
\end{equation}

The raw prediction $\tilde{y}_{i,j}^{\,p_t}$ is subsequently processed by a validator function $\mathit{V}(\cdot)$ that checks for adherence of predictions to a predefined output schema customized for DLA and KIE. Invalid inferences are excluded from evaluation.
\begin{equation}
    \hat{y}_{i,j}^{\,p_t} = \mathit{V}\!\left(\tilde{y}_{i,j}^{\,p_t}\right)
\end{equation}

The validated prediction is then compared against the ground-truth annotation $y_{i,j}$ to obtain page-level performance measurements aggregated across pages and forms. For each predicted form entity, the output also contains a brief label justification and a confidence score for the bounding box coordinates.

For DLA, we define a prompt set as 
$\mathit{P_{DLA}} = \left\{ p_{\text{zs}},\; p_{\text{cot}} \right\}$,
where $p_{\text{zs}}$ and $p_{\text{cot}}$ denote DLA-specific zero-shot and chain-of-thought prompts, respectively. For KIE the prompt set $\mathit{P_{KIE}}$ contains a zero-shot evaluation prompt. To observe effects of entity set size, in addition to the granular form entity set containing 26 entity categories, we developed a minimal set containing 5 coarse entity categories and mapped all original 26 categories to coarse categories (See \cref{tab:layout_categories}).
    
    \section{Results}
    \subsection{Document Layout Analysis (DLA)}

\begin{table*}[!htb]
    \centering
    \caption{
        Layout analysis performance for both Granular and Coarse label sets. \textbf{Bold} indicates the best overall result per column within each entity set, and \textcolor{blue!60}{Blue} indicates the best result within each combination of prompt variant (Zero-Shot: ZS, Chain-of-Thoughts: CoT) and reasoning effort (Low, High). $\mu$IoU is used to show overall average IoU regions for all predicted bounding boxes. $\mu$IoU$@$[0.3, 0.5] is used to show average of bounding boxes that are over the thresholds 0.3 and 0.5, the thresholds are selected based on works from Form-NLU \cite{formnlu} and OmniDocBench\cite{omnidocbench}. Across all prompt, reasoning effort, and entity set granularity level combination, \model{Gemini 3 Pro} outperforms \model{GPT-5.2} and \model{Claude Opus 4.6}. For each metric higher ($\uparrow$) is better.
    }
    \label{tab:dla_combined}
    \resizebox{0.8\textwidth}{!}{%
        \setlength{\tabcolsep}{6pt}
        
        \begin{tabular}{ccl cccc cccc}
            \toprule
            \multirow{2}{*}{\makecell{\textbf{Reasoning}\\\textbf{Effort}}} &
            \multirow{2}{*}{\textbf{Prompt}} &
            \multirow{2}{*}{\textbf{Model}} & 
            \multicolumn{4}{c}{\textbf{IoU$@$0.3}} & 
            \multicolumn{4}{c}{\textbf{IoU$@$0.5}} \\
            \cmidrule(lr){4-7} \cmidrule(lr){8-11}
            & & &
            \textbf{mAP} & \textbf{F1} & \textbf{$\mu$IoU} & \textbf{$\mu$IoU$@$0.3} & 
            \textbf{mAP} & \textbf{F1} & \textbf{$\mu$IoU} & \textbf{$\mu$IoU$@$0.5} \\

            \midrule
            \multicolumn{11}{c}{\textbf{Granular Entity Set}} \\
            \midrule
            
            \multirow{10}{*}{Low}
            & \multirow{5}{*}{ZS}
            & GPT-5.2
            & 0.0530 & 0.1158 & 0.1242 & 0.4556
            & 0.0155 & 0.0444 & 0.1259 & 0.6175 \\
            & & Claude Opus 4.6
            & 0.0041 & 0.0161 & 0.0316 & 0.4334
            & 0.0010 & 0.0049 & 0.0323 & 0.5998 \\
            & & Gemini-3 Pro
            & \textcolor{blue!60}{0.0900} & \textcolor{blue!60}{0.1953} & \textbf{0.2430} & \textbf{0.5647}
            & \textcolor{blue!60}{0.0483} & \textcolor{blue!60}{0.1222} & \textbf{0.2449} & \textbf{0.6787} \\
            & & Kimi K2.5
            & 0.0156 & 0.0566 & 0.0676 & 0.4787
            & 0.0048 & 0.0196 & 0.0695 & 0.6087 \\
            & & Qwen 3.6-Plus
            & 0.0097 & 0.0422 & 0.0691 & 0.4184
            & 0.0008 & 0.0088 & 0.0705 & 0.6048 \\
            
            \cmidrule(lr){2-11}
            
            & \multirow{5}{*}{CoT}
            & GPT-5.2
            & 0.0643 & 0.1273 & 0.1343 & 0.4782
            & 0.0266 & 0.0618 & 0.1349 & 0.5991 \\
            & & Claude Opus 4.6
            & 0.0037 & 0.0174 & 0.0355 & 0.4206
            & 0.0007 & 0.0051 & 0.0363 & 0.5907 \\
            & & Gemini-3 Pro
            & \textcolor{blue!60}{0.0853} & \textcolor{blue!60}{0.1943} & \textcolor{blue!60}{0.2420} & \textcolor{blue!60}{0.5390}
            & \textcolor{blue!60}{0.0452} & \textcolor{blue!60}{0.1207} & \textcolor{blue!60}{0.2433} & \textcolor{blue!60}{0.6594} \\
            & & Kimi K2.5
            & 0.0105 & 0.0480 & 0.0690 & 0.4530
            & 0.0027 & 0.0182 & 0.0707 & 0.5989 \\
            & & Qwen 3.6-Plus
            & 0.0083 & 0.0371 & 0.0650 & 0.4101
            & 0.0010 & 0.0072 & 0.0717 & 0.6025 \\
            
            \midrule
            
            \multirow{10}{*}{High}
            & \multirow{5}{*}{ZS}
            & GPT-5.2
            & 0.0511 & 0.1192 & 0.1148 & 0.4658
            & 0.0157 & 0.0481 & 0.1157 & 0.6134 \\
            & & Claude Opus 4.6
            & 0.0074 & 0.0321 & 0.0460 & 0.4408
            & 0.0012 & 0.0075 & 0.0465 & 0.5771 \\
            & & Gemini-3 Pro
            & \textbf{0.1177} & \textbf{0.2282} & \textcolor{blue!60}{0.2243} & \textcolor{blue!60}{0.5312}
            & \textcolor{blue!60}{0.0641} & \textcolor{blue!60}{0.1411} & \textcolor{blue!60}{0.2257} & \textcolor{blue!60}{0.6419} \\
            & & Kimi K2.5
            & 0.0453 & 0.1117 & 0.1132 & 0.4617
            & 0.0150 & 0.0441 & 0.1140 & 0.6123 \\
            & & Qwen 3.6-Plus
            & 0.0175 & 0.0606 & 0.0661 & 0.4501
            & 0.0040 & 0.0234 & 0.0668 & 0.5910 \\
            
            \cmidrule(lr){2-11}
            
            & \multirow{5}{*}{CoT}
            & GPT-5.2
            & 0.0615 & 0.1291 & 0.1234 & 0.4723
            & 0.0197 & 0.0581 & 0.1247 & 0.5985 \\
            & & Claude Opus 4.6
            & 0.0065 & 0.0310 & 0.0501 & 0.4350
            & 0.0016 & 0.0104 & 0.0505 & 0.5683 \\
            & & Gemini-3 Pro
            & \textcolor{blue!60}{0.1134} & \textcolor{blue!60}{0.2206} & \textcolor{blue!60}{0.2193} & \textcolor{blue!60}{0.5438}
            & \textbf{0.0643} & \textbf{0.1435} & \textcolor{blue!60}{0.2205} & \textcolor{blue!60}{0.6586} \\
            & & Kimi K2.5
            & 0.0476 & 0.1133 & 0.1162 & 0.4559
            & 0.0152 & 0.0435 & 0.1175 & 0.6130 \\
            & & Qwen 3.6-Plus
            & 0.0166 & 0.0609 & 0.0657 & 0.4329
            & 0.0052 & 0.0216 & 0.0665 & 0.5959 \\
            
            \midrule
            \multicolumn{11}{c}{\textbf{Coarse Entity Set}} \\
            \midrule
            
            \multirow{10}{*}{Low}
            & \multirow{5}{*}{ZS}
            & GPT-5.2
            & 0.1800 & 0.2974 & 0.2742 & 0.5177
            & 0.0852 & 0.1654 & 0.2767 & 0.6495 \\
            & & Claude Opus 4.6
            & 0.0168 & 0.0555 & 0.0697 & 0.4081
            & 0.0042 & 0.0138 & 0.0722 & 0.5875 \\
            & & Gemini-3 Pro
            & \textcolor{blue!60}{0.2578} & \textcolor{blue!60}{0.4020} & \textcolor{blue!60}{0.3602} & \textcolor{blue!60}{0.6066}
            & \textbf{0.1606} & \textcolor{blue!60}{0.2847} & \textcolor{blue!60}{0.3616} & \textcolor{blue!60}{0.6986} \\
            & & Kimi K2.5
            & 0.0469 & 0.1213 & 0.1306 & 0.4529
            & 0.0131 & 0.0392 & 0.1344 & 0.6433 \\
            & & Qwen 3.6-Plus
            & 0.0330 & 0.1273 & 0.1511 & 0.4111
            & 0.0021 & 0.0296 & 0.1584 & 0.5865 \\
            
            \cmidrule(lr){2-11}
            
            & \multirow{5}{*}{CoT}
            & GPT-5.2
            & 0.1671 & 0.2861 & 0.2659 & 0.5174
            & 0.0781 & 0.1571 & 0.2682 & 0.6409 \\
            & & Claude Opus 4.6
            & 0.0131 & 0.0578 & 0.0731 & 0.4164
            & 0.0022 & 0.0132 & 0.0751 & 0.5884 \\
            & & Gemini-3 Pro
            & \textcolor{blue!60}{0.2513} & \textcolor{blue!60}{0.4051} & \textcolor{blue!60}{0.3682} & \textcolor{blue!60}{0.6085}
            & \textcolor{blue!60}{0.1440} & \textcolor{blue!60}{0.2827} & \textcolor{blue!60}{0.3698} & \textcolor{blue!60}{0.7002} \\
            & & Kimi K2.5
            & 0.0426 & 0.1293 & 0.1342 & 0.4385
            & 0.0057 & 0.0359 & 0.1398 & 0.6033 \\
            & & Qwen 3.6-Plus
            & 0.0377 & 0.1199 & 0.1512 & 0.4123
            & 0.0021 & 0.0250 & 0.1577 & 0.5981 \\
            
            \midrule
            
            \multirow{10}{*}{High}
            & \multirow{5}{*}{ZS}
            & GPT-5.2
            & 0.1434 & 0.2866 & 0.2561 & 0.5046
            & 0.0532 & 0.1391 & 0.2592 & 0.6472 \\
            & & Claude Opus 4.6
            & 0.0440 & 0.1168 & 0.1181 & 0.4283
            & 0.0136 & 0.0454 & 0.1199 & 0.5932 \\
            & & Gemini-3 Pro
            & \textbf{0.2646} & \textbf{0.4116} & \textbf{0.3729} & \textbf{0.6125}
            & \textcolor{blue!60}{0.1480} & \textbf{0.2915} & \textbf{0.3745} & \textbf{0.7003} \\
            & & Kimi K2.5
            & 0.1336 & 0.2638 & 0.2259 & 0.4748
            & 0.0466 & 0.1190 & 0.2283 & 0.6205 \\
            & & Qwen 3.6-Plus
            & 0.0589 & 0.1692 & 0.1583 & 0.4740
            & 0.0204 & 0.0714 & 0.1604 & 0.6143 \\
            
            \cmidrule(lr){2-11}
            
            & \multirow{5}{*}{CoT}
            & GPT-5.2
            & 0.1354 & 0.2776 & 0.2549 & 0.5158
            & 0.0663 & 0.1546 & 0.2575 & 0.6340 \\
            & & Claude Opus 4.6
            & 0.0358 & 0.1012 & 0.1112 & 0.4518
            & 0.0091 & 0.0385 & 0.1132 & 0.5829 \\
            & & Gemini-3 Pro
            & \textcolor{blue!60}{0.2444} & \textcolor{blue!60}{0.3928} & \textcolor{blue!60}{0.3499} & \textcolor{blue!60}{0.6071}
            & \textcolor{blue!60}{0.1426} & \textcolor{blue!60}{0.2751} & \textcolor{blue!60}{0.3511} & \textcolor{blue!60}{0.6975} \\
            & & Kimi K2.5
            & 0.1361 & 0.2602 & 0.2245 & 0.4874
            & 0.0472 & 0.1217 & 0.2272 & 0.6281 \\
            & & Qwen 3.6-Plus
            & 0.0693 & 0.1749 & 0.1615 & 0.4766 
            & 0.0200 & 0.0733 & 0.1637 & 0.6218 \\
            
            \bottomrule
        \end{tabular}%
    }
\end{table*}
\begin{table*}[htb]
    \centering
    \caption{Bangla and English Form Comparison for DLA (IoU@0.3) under different experimental setups. LR and HR stand for Low Reasoning and High Reasoning respectively. ZS and CoT denote Zero-Shot and Chain-of-Thought prompts. Shapes indicate higher performance for Bangla (\ban), English (\eng), or neither (\eqv) language. For granular entity set, the effect of language is minimal ($\le 0.02$). For coarse entity set, effect of language is slightly more with max$\Delta$ of $~0.12$. For each metric higher ($\uparrow$) is better.}
    \label{tab:bn_en_combined}

    \resizebox{0.65\textwidth}{!}{
        \setlength{\tabcolsep}{8pt}
        \begin{tabular}{@{}llcccccc@{}}
            \toprule
            \multirow{2}{*}{\textbf{Variant}} &
            \multirow{2}{*}{\textbf{Model}} &
            \multicolumn{2}{c}{\textbf{Bangla}} &
            \multicolumn{2}{c}{\textbf{English}} &
            \multicolumn{2}{c}{\textbf{Difference (Bn-En)}}\\
            \cmidrule(lr){3-4}\cmidrule(lr){5-6}\cmidrule(lr){7-8}
            & & \textbf{mAP} & \textbf{F1} & \textbf{mAP} & \textbf{F1} & \textbf{$\Delta$ mAP} & \textbf{$\Delta$ F1} \\
            \midrule
            
            \multicolumn{8}{c}{\textbf{Granular Entity Set}}\\
            \midrule
            
            \multirow{2}{*}{\shortstack{LR + ZS}}
            & GPT-5.2 & 0.09 & 0.16 & 0.09 & 0.14 & \eqv\;0.00 & \ban\;0.02 \\
            & Gemini 3 Pro & 0.16 & 0.24 & 0.15 & 0.24 & \ban\;0.01 & \eqv\;0.00 \\
            \midrule
            
            \multirow{2}{*}{\shortstack{LR + CoT}}
            & GPT-5.2 & 0.12 & 0.18 & 0.11 & 0.15 & \ban\;0.01 & \ban\;0.03 \\
            & Gemini 3 Pro & 0.15 & 0.21 & 0.14 & 0.23 & \ban\;0.01 & \eng\;0.02 \\
            \midrule
            
            \multirow{2}{*}{\shortstack{HR + ZS}}
            & GPT-5.2 & 0.09 & 0.16 & 0.11 & 0.15 & \eng\;0.02 & \ban\;0.01\\
            & Gemini 3 Pro & 0.16 & 0.24 & 0.16 & 0.23 & \eqv\;0.00 & \ban\;0.01\\
            \midrule
            
            \multirow{2}{*}{\shortstack{HR + CoT}}
            & GPT-5.2 & 0.12 & 0.18 & 0.13 & 0.17 & \eng\;0.01 & \ban\;0.01 \\
            & Gemini 3 Pro & 0.15 & 0.21 & 0.16 & 0.24 & \eng\;0.01 & \eng\;0.03 \\
            
            \midrule
            \multicolumn{8}{c}{\textbf{Coarse Entity Set}}\\
            \midrule
            
            \multirow{2}{*}{\shortstack{LR + ZS}}
            & GPT-5.2 & 0.34 & 0.44 & 0.22 & 0.32 & \ban\;0.12 & \ban\;0.12 \\
            & Gemini 3 Pro & 0.38 & 0.51 & 0.35 & 0.42 & \ban\;0.03 & \ban\;0.09\\
            \midrule
            
            \multirow{2}{*}{\shortstack{LR + CoT}}
            & GPT-5.2 & 0.29 & 0.44 & 0.28 & 0.40 & \ban\;0.01 & \ban\;0.04 \\
            & Gemini 3 Pro & 0.35 & 0.47 & 0.25 & 0.40 & \ban\;0.09 & \ban\;0.07\\
            \midrule
            
            \multirow{2}{*}{\shortstack{HR + ZS}}
            & GPT-5.2 & 0.34 & 0.44 & 0.27 & 0.35 & \ban\;0.07 & \ban\;0.09\\
            & Gemini 3 Pro & 0.38 & 0.51 & 0.36 & 0.47 & \ban\;0.02 & \ban\;0.04\\
            \midrule
            
            \multirow{2}{*}{\shortstack{HR + CoT}}
            & GPT-5.2 & 0.29 & 0.44 & 0.27 & 0.38 & \ban\;0.02 & \ban\;0.07\\
            & Gemini 3 Pro & 0.35 & 0.47 & 0.42 & 0.52 & \eng\;0.07 & \eng\;0.05\\
            
            \bottomrule
        \end{tabular}
    }
\end{table*}

\cref{tab:dla_combined} reports DLA performance across models, prompting strategies, reasoning effort levels, and entity granularities. Overall, \model{Gemini 3 Pro} achieves the best results in most configurations, followed by \model{GPT-5.2}, while \model{Claude Opus 4.6} is consistently the weakest. Performance improves noticeably from the granular to the coarse entity set. For instance, under high reasoning effort with zero-shot prompting, \model{Gemini 3 Pro} increases from 0.1177 mAP on the granular entity set to 0.2646 on the coarse set at IoU@0.3, with similar trends observed across other models. Among the open-source models, \model{Kimi K2.5} and \model{Qwen 3.6-Plus} trail the proprietary leaders under low reasoning effort. However, \model{Kimi K2.5} benefits substantially from increased reasoning effort: its coarse mAP@0.3 under zero-shot prompting rises from 0.0469 to 0.1336, approaching \model{GPT-5.2} (0.1434), whereas \model{Qwen 3.6-Plus} improves only modestly, from 0.0330 to 0.0589.

Increasing reasoning effort has mixed effects, yielding small gains in some settings but slight decreases in Avg.\ IoU in others. CoT prompting likewise provides limited benefit, often performing comparably to zero-shot prompting.

The effect of language on MLLM performance is minimal. \cref{tab:bn_en_combined} compares Bangla and English performance for the two strongest models, \model{Gemini 3 Pro} and \model{GPT-5.2}. For both models, the two languages yield similar results across most configurations, with $\Delta$mAP typically within 0.02 on the granular entity set. Overall, DLA performance remains largely consistent across both languages and relatively low for granular entities.

\subsection{Key Information Extraction (KIE)}

\cref{tab:kie_eval_bn_en} reports KIE performance using Precision, Recall, F1, and Normalized Edit Similarity (NES). All models score substantially higher here than on DLA. Among proprietary models, \model{Gemini 3 Pro} leads on Bangla (F1 0.848, NES 0.866), while \model{GPT-5.2} leads on English (F1 0.847, NES 0.851). \model{Claude Opus 4.6} trails both but remains competitive, unlike in DLA, where it was the weakest performer. Among open-source models, \model{Qwen 3.6-Plus} performs best on both languages (Bangla F1 0.794, English F1 0.835), ahead of \model{Kimi K2.5}. It ranks second overall on Bangla, surpassing \model{GPT-5.2} (F1 0.781) and \model{Claude Opus 4.6} (F1 0.678), and matches \model{Claude Opus 4.6} on English (F1 0.835), narrowing the gap between open-source and proprietary models.

Unlike DLA, where language differences were minimal, KIE varies more clearly by language. Most models score higher on English-the gap is largest for \model{Claude Opus 4.6} (0.678 vs.\ 0.835 F1) and \model{Kimi K2.5} (0.681 vs.\ 0.828), while only \model{Gemini 3 Pro} favors Bangla (0.848 vs.\ 0.828). Consequently, the best model is language-dependent: \model{Gemini 3 Pro} on Bangla and \model{GPT-5.2} on English. So cross-lingual differences are stronger in KIE compared to layout detection.

\begin{table}[bth]
    \centering                 
    \caption{KIE evaluation results on Bangla (BN) and English (EN) forms. Normalized Edit Similarity (NES) is reported along with Precision, Recall, and Macro-F1. English forms are selected from the same domain as Bangla. \model{Gemini 3 Pro} performs the best for Bangla forms, whereas for English \model{GPT-5.2} performs better.}
    \label{tab:kie_eval_bn_en}

    \resizebox{0.65\textwidth}{!}{%
        \setlength{\tabcolsep}{6pt}
        \begin{tabular}{lcccccccc}
            \toprule
            & \multicolumn{2}{c}{\textbf{Precision}}
            & \multicolumn{2}{c}{\textbf{Recall}}
            & \multicolumn{2}{c}{\textbf{Macro-F1}}
            & \multicolumn{2}{c}{\textbf{NES}} \\
            \cmidrule(lr){2-3}\cmidrule(lr){4-5}\cmidrule(lr){6-7}\cmidrule(lr){8-9}
            \textbf{Model} & \textbf{BN} & \textbf{EN} & \textbf{BN} & \textbf{EN} & \textbf{BN} & \textbf{EN} & \textbf{BN} & \textbf{EN} \\
            \midrule
            
            \multicolumn{9}{c}{\textbf{Proprietary}} \\
            \cmidrule{1-9}
            
            GPT-5.2
            & 0.784 & \textbf{0.852}
            & 0.780 & \textbf{0.844}
            & 0.781 & \textbf{0.847}
            & 0.796 & \textbf{0.851} \\
            
            {Claude Opus 4.6}
            & 0.675 & 0.839
            & 0.685 & 0.832
            & 0.678 & 0.835
            & 0.704 & 0.840 \\
            
            Gemini 3 Pro
            & \textbf{0.844} & 0.832
            & \textbf{0.853} & 0.827
            & \textbf{0.848} & 0.828
            & \textbf{0.866} & 0.833 \\
            
            \midrule
            \multicolumn{9}{c}{\textbf{Open-Source}} \\
            \cmidrule{1-9}
            
            Kimi K2.5
            & 0.684 & 0.835
            & 0.681 & 0.825
            & 0.681 & 0.828
            & 0.700 & 0.832 \\
            
            Qwen 3.6-Plus
            & 0.798 & 0.839
            & 0.793 & 0.834
            & 0.794 & 0.835
            & 0.804 & 0.844 \\
            \bottomrule
        \end{tabular}%
    }
\end{table}

% can be added later to check intra-qwen perofrmance bump
\begin{comment}
Qwen 3.5
& 0.773 & 0.794
& 0.764 & 0.791
& 0.767 & 0.792
& 0.774 & 0.805 \\
\end{comment}

    \section{Discussion}
    \textbf{Granularity of form entities strongly affects DLA performance}. Across all models, performance improves markedly when coarse entity set is used. Reducing the number of entity types substantially increases both mAP and F1 across most configurations. For example, mAP score for \model{Gemini-3 Pro} improves from 0.1177 to 0.2646, for granular to coarse entity set shift for IoU@0.3, more than doubling performance. This indicates that much of the challenge in DLA lies in distinguishing fine-grained entity types; when grouped into broader categories, models localize layout regions more reliably.

\textbf{Increasing reasoning effort does not consistently improve geometric precision}. Across both entity sets and prompting strategies, higher reasoning effort yields only small and inconsistent performance changes. In several cases, it slightly reduces localization quality, particularly Avg.\ IoU. For example, \model{Gemini-3 Pro} shows lower Avg.\ IoU when moving from low to high reasoning effort in the granular setting, while \model{GPT-5.2} shows small improvements in some configurations. Overall, this mixed behavior suggests that additional reasoning does not reliably improve geometric alignment; layout prediction appears to depend more on visual-spatial pattern recognition than extended reasoning.

\textbf{Chain-of-thought prompting provides limited gains for DLA}. Across most models and settings, CoT prompting does not clearly outperform zero-shot prompting; performance often remains similar or slightly declines in mAP and F1. For example, \model{Gemini-3 Pro} achieves slightly lower mAP with CoT than with zero-shot in multiple settings across both entity sets. This suggests that encouraging multi-step reasoning offers limited benefit for DLA, which appears to rely more on visual-spatial understanding than explicit reasoning chains.

\textbf{Language has limited impact on DLA performance}. Across most experimental variants, performance differences between Bangla and English forms are small, with negligible gaps in mAP and F1 (see \cref{tab:bn_en_combined}). In the granular entity setting, differences are near zero (e.g., $\Delta$mAP $\leq 0.02$ across most setups). Even in the coarse setting, improvements are inconsistent and modest relative to overall performance. This suggests that DLA difficulty for MLLMs stems mainly from geometric localization and structural understanding rather than language.

\textbf{MLLMs perform substantially better on KIE than on DLA}. Across all models, KIE results are considerably stronger than document layout analysis performance. The best-performing model achieves F1 scores above 0.84 on Bangla forms and 0.80 on English forms (see \cref{tab:kie_eval_bn_en}), while DLA results remain much lower across comparable configurations. This indicates that MLLMs are more effective at textual understanding and semantic extraction than at precise geometric localization of layout regions.

\textbf{Language differences are more pronounced in KIE than in DLA}. Unlike DLA experiments-where Bangla and English forms yield nearly identical performance. KIE results show clearer language-dependent variation across models. For example, \model{Gemini 3 Pro} achieves the best performance on Bangla forms (F1 = 0.848), while \model{GPT-5.2} performs best on English forms (F1 = 0.800) (see \cref{tab:kie_eval_bn_en}). This suggests that language-specific factors affect text-heavy extraction tasks, while layout detection remains largely language-agnostic.

    \section{Conclusion}
    We propose \textbf{BaFCo}, the first Bangla document understanding dataset featuring well-curated multi-domain Bangladeshi government forms. It includes an expanded, fine-grained set of 26 form entities designed for better Bangla document comprehension. We evaluated flagship MLLMs on DLA and KIE using both granular and coarse entity sets, revealing significant performance gaps. DLA performance is inconsistent across models and sensitive to reasoning effort and prompting strategies, whereas KIE results are generally stronger, though some lexical mismatches remain.

    \section{Limitations and Future Works}
    The scope of generative model-based Bangla form comprehension can be expanded along several directions:

(a) Our experiments focus on DLA and KIE, key building blocks for downstream tasks like summarization and ontology generation. Future work will explore frontier models on broader document comprehension tasks.

(b) Low-resource languages face greater challenges in curating high-quality domain-specific documents due to limited data access, lack of experts, and other constraints. Although BaFCo contains fewer annotated forms than high-resource datasets like \cite{omnidocbench, publaynet, xfund}, it aims to stimulate research on Bangla form comprehension. Future work will expand both the quantity and diversity of forms.

(c) To address model limitations and scarcity of large-scale Bangla data,  lightweight post-training methods and agentic AI approaches can be explored.

    \section*{Acknowledgement}
    This work was partially funded by Independent University, Bangladesh (IUB).

    % \balance
    % \clearpage
    \bibliographystyle{splncs04}
    \bibliography{references}
    
    \clearpage
    \appendix
    \section*{Appendix}
    % \subsection{Reproducibility}
% Our evaluation code and dataset annotations are available at the following anonymized link: \url{https://figshare.com/s/5230f186bb3e7fc98373}

\section{Form Entity Distribution}
\label{appendix:entity_distribution}

\cref{fig:entity_dist_charts} visualizes the ground-truth entity distribution under the coarse (5-class) and granular (26-class) label sets for DLA, with exact counts and percentages listed in \cref{tab:entity_dist}. The distribution is strongly long-tailed, reflecting the natural composition of real-world Bangladeshi forms. At the coarse level, \emph{Fields} dominate (64.6\%), followed by \emph{Table} (27.3\%), with \emph{Others} (3.9\%), \emph{Headings} (3.8\%), and \emph{Image} (0.4\%) forming the tail. A similar pattern persists at the granular level: \emph{Form Key} and \emph{Form Value} together account for roughly half of all entities, while table-cell classes comprise most of the remainder. Rare classes such as \emph{Table Section Title} and \emph{Others}, by contrast, occur only a handful of times.

\begin{figure*}[!htb]
      \centering
      \begin{subfigure}{\textwidth}
          \centering
          \includegraphics[width=\textwidth]{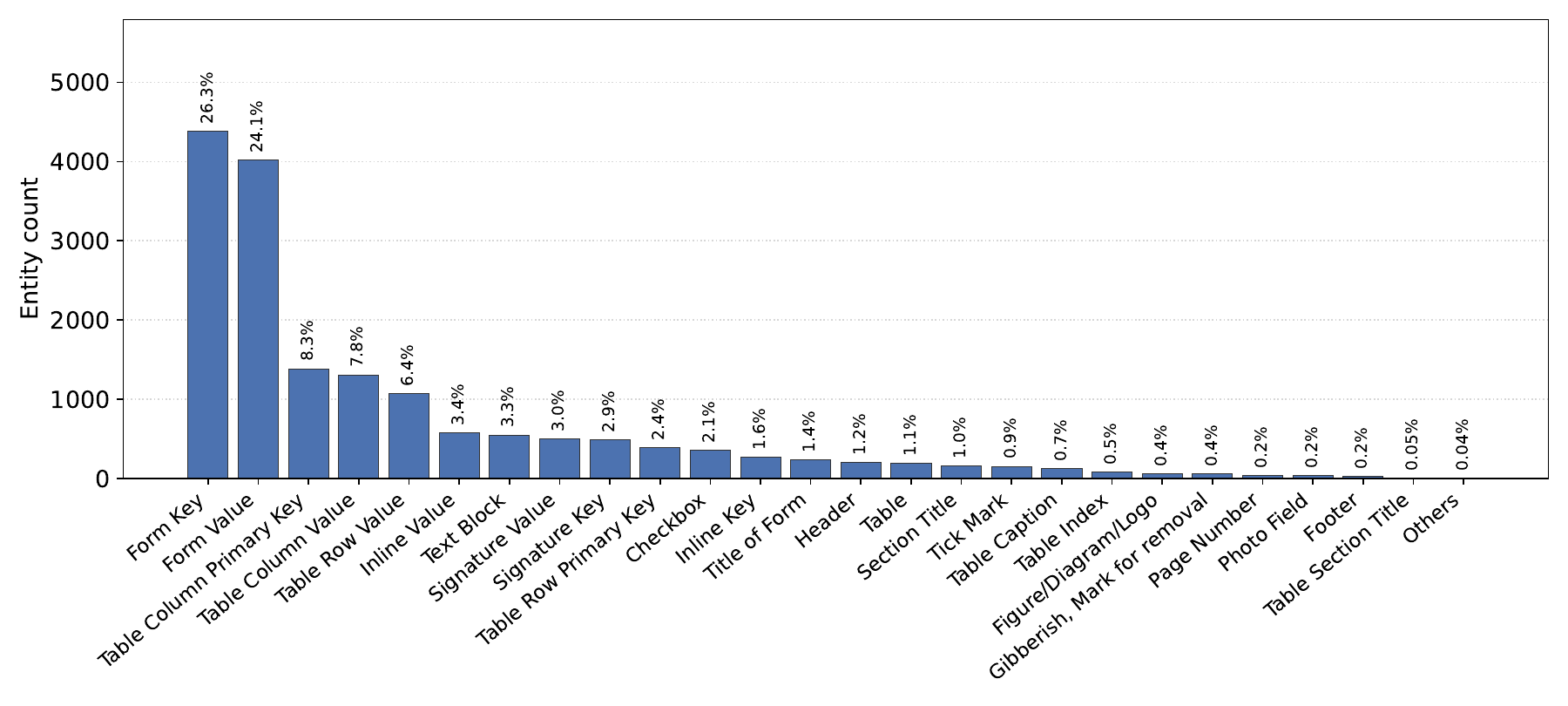}
          \caption{Granular entity set}
          \label{fig:entity_gran}
      \end{subfigure}
      \\[8pt]
      \begin{subfigure}{0.7\textwidth}
          \centering
          \includegraphics[width=0.85\textwidth]{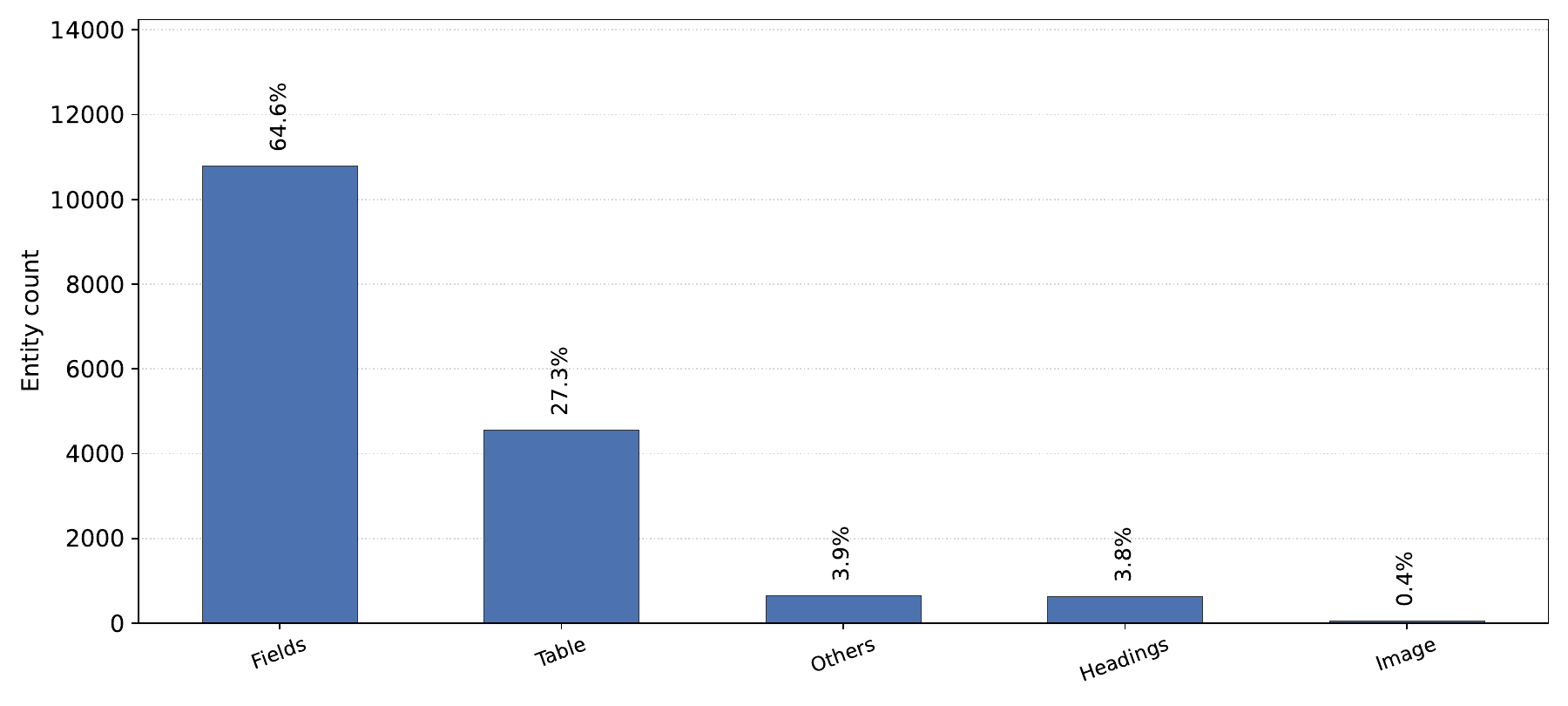}
          \caption{Coarse entity set}
          \label{fig:entity_coarse}
      \end{subfigure}
      \caption{DLA entity distributions over the annotated forms, shown for the (a)~granular (26-label) and (b)~coarse (5-label) entity sets. Both follow a long-tailed pattern dominated by form key/value fields. Exact counts are given in \cref{tab:entity_dist}.}
      \label{fig:entity_dist_charts}
  \end{figure*}
\begin{table*}[!htb]
      \centering
      \caption{DLA entity distribution over the annotated forms. Granular labels are grouped under their coarse category; \textbf{bold} rows give the coarse
  subtotals.}
      \label{tab:entity_dist}
      \resizebox{0.8\textwidth}{!}{%
          \setlength{\tabcolsep}{6pt}
          \begin{tabular}{lrr @{\hspace{2em}} lrr}
              \toprule
              \textbf{Entity} & \textbf{Count} & \textbf{\%} & \textbf{Entity} & \textbf{Count} & \textbf{\%} \\
              \midrule
              \textbf{Fields} & \textbf{10{,}790} & \textbf{64.6} & \textbf{Table} & \textbf{4{,}563} & \textbf{27.3} \\
              \quad Form Key & 4{,}389 & 26.3 & \quad Table Column Primary Key & 1{,}385 & 8.3 \\
              \quad Form Value & 4{,}026 & 24.1 & \quad Table Column Value & 1{,}308 & 7.8 \\
              \quad Inline Value & 573 & 3.4 & \quad Table Row Value & 1{,}073 & 6.4 \\
              \quad Signature Value & 496 & 3.0 & \quad Table Row Primary Key & 393 & 2.4 \\
              \quad Signature Key & 490 & 2.9 & \quad Table & 190 & 1.1 \\
              \quad Checkbox & 357 & 2.1 & \quad Table Caption & 122 & 0.7 \\
              \quad Inline Key & 270 & 1.6 & \quad Table Index & 84 & 0.5 \\
              \quad Tick Mark & 152 & 0.9 & \quad Table Section Title & 8 & 0.05 \\
              \quad Photo Field & 37 & 0.2 & \textbf{Headings} & \textbf{626} & \textbf{3.8} \\
              \textbf{Others} & \textbf{653} & \textbf{3.9} & \quad Title of Form & 235 & 1.4 \\
              \quad Text Block & 548 & 3.3 & \quad Header & 203 & 1.2 \\
              \quad Gibberish, Mark for removal & 60 & 0.4 & \quad Section Title & 160 & 1.0 \\
              \quad Page Number & 39 & 0.2 & \quad Footer & 28 & 0.2 \\
              \quad Others & 6 & 0.04 & \textbf{Image} & \textbf{61} & \textbf{0.4} \\
               & &  & \quad Figure/Diagram/Logo & 61 & 0.4 \\
              \midrule
              \multicolumn{6}{c}{\textbf{Total: 16{,}693 entities \;(100.0\%)}} \\
              \bottomrule
          \end{tabular}%
      }
  \end{table*}

\section{Form Difficulty Examples}

\cref{fig:difficulty_examples} shows one representative form per difficulty tier. The Easy example contains only key--value fields and headings, with no tables or checkboxes. The Medium example contains a columns-only table alongside key--value fields. The Hard example contains a full row--column table with multiple pre-defined rows, characteristic of the densest forms in the dataset.

\begin{figure*}[!htb]
    \centering
    
    \begin{subfigure}{0.31\textwidth}\centering
        \includegraphics[width=\linewidth]{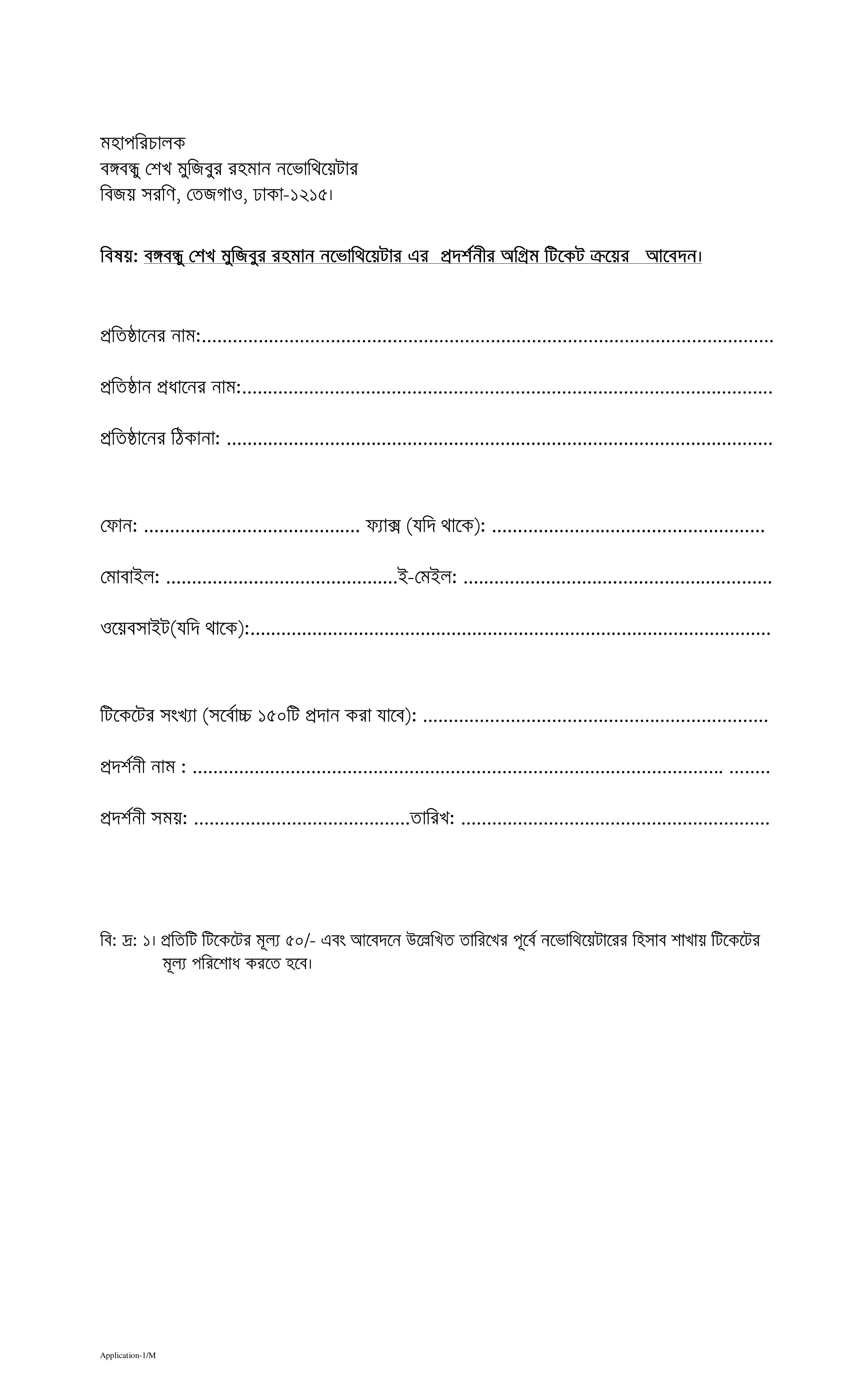}
        \caption{Easy}\label{fig:diff_easy}
    \end{subfigure}\hfill
    \begin{subfigure}{0.31\textwidth}\centering
        \includegraphics[width=\linewidth]{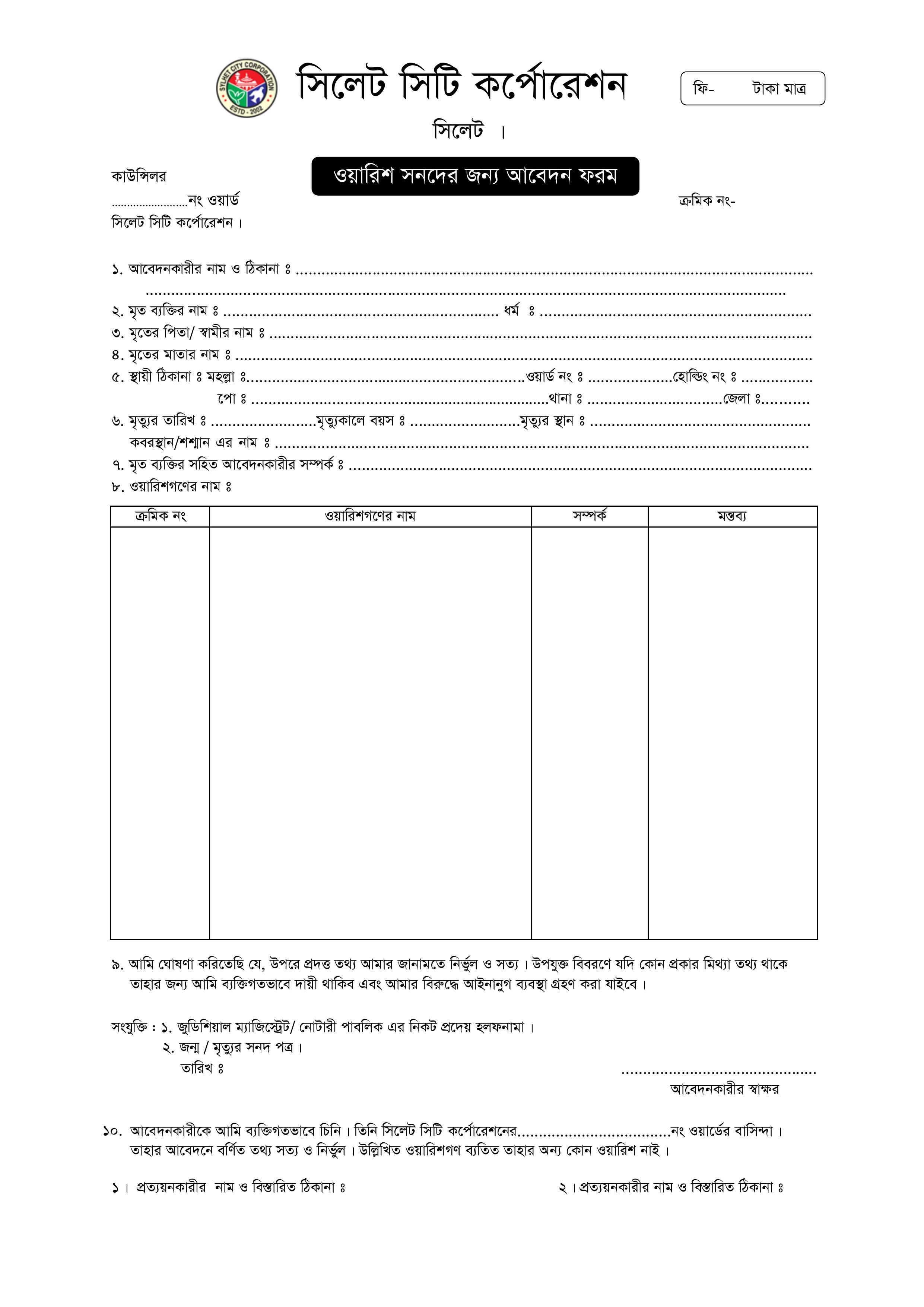}
        \caption{Medium}\label{fig:diff_medium}
    \end{subfigure}\hfill
    \begin{subfigure}{0.31\textwidth}\centering
        \includegraphics[width=\linewidth]{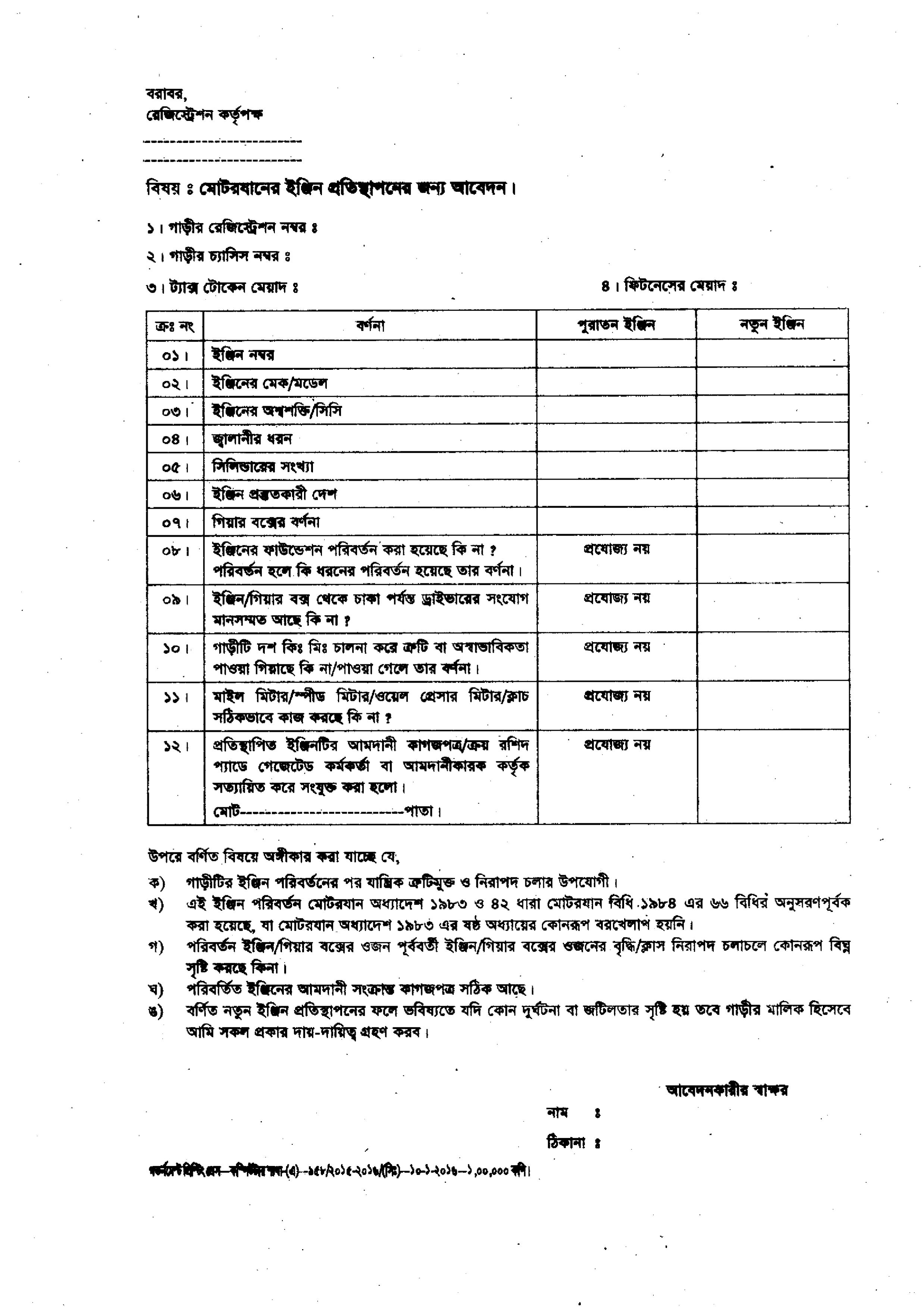}
        \caption{Hard}\label{fig:diff_hard}
    \end{subfigure}
    \caption{
        Representative forms for each difficulty tier. 
        (a)~Easy forms contain only key--value fields and headings; (b)~Medium forms add a columns-only table (here, the list of heirs); 
        (c)~Hard forms contain a full row--column table with pre-defined rows.
    }
    \label{fig:difficulty_examples}
  \end{figure*}

\section{Qualitative Examples}

We manually inspect predictions from all evaluated models to identify common failure modes. This section summarizes recurring patterns and presents representative examples.

\subsection{Document Layout Analysis (DLA)}

\cref{fig:all_granular} illustrates representative success and failure cases for MLLM predictions on \emph{granular} form entity categories.

\textbf{Spatial misalignment:} A common failure occurs when models correctly classify entities but localize them inaccurately. As shown in \cref{fig:gran_a}, predicted bounding boxes are systematically shifted toward the upper portion of the form despite correct category assignments.

\textbf{Entity miscategorization:} Another frequent failure is semantic rather than spatial. In \cref{fig:gran_b}, bounding box localization is largely correct, but key--value entities are assigned incorrect labels, suggesting that spatial localization and semantic classification are only weakly coupled.

\textbf{Hallucination:} The most severe failure mode is shown in \cref{fig:gran_c}, where the model confidently predicts entities in visually empty regions. In particular, it hallucinates inline keys in the blank top-left and top-right areas, a tendency that becomes more pronounced in denser forms.

\textbf{Successful prediction:} In contrast, \cref{fig:gran_d} shows a successful prediction. The sparse layout reduces ambiguity, allowing accurate localization and categorization across the form.

\cref{fig:all_coarse} presents analogous examples for the \emph{coarse} entity taxonomy. Although reducing the label space improves overall localization, the same failure modes remain persist in more subtle forms.

\textbf{Positional bias:} As shown in \cref{fig:coarse_a} , predictions are often concentrated in the upper portion of the form, with much of the lower half omitted. This indicates a positional bias, potentially arising from how MLLMs process long visual documents.

\textbf{Failure of hierarchical decomposition:} \cref{fig:coarse_b} illustrates a complementary limitation: even when a large enclosing region is detected, nested sub-fields are frequently missed entirely. This suggests that current MLLMs reason at a coarse spatial granularity and struggle with hierarchical region decomposition.

\textbf{Successful predictions:} \cref{fig:coarse_c} and \cref{fig:coarse_d} show that structured, low-density forms and visually distinctive layouts are more tractable. \model{GPT-5.2} performs well on the simpler form in \cref{fig:coarse_c}, while \model{Gemini 3 Pro} accurately localizes entities even in the more complex example shown in \cref{fig:coarse_d}.

\begin{figure}[!htb]
    \centering
    \begin{subfigure}{0.45\textwidth}
        \includegraphics[width=\linewidth]{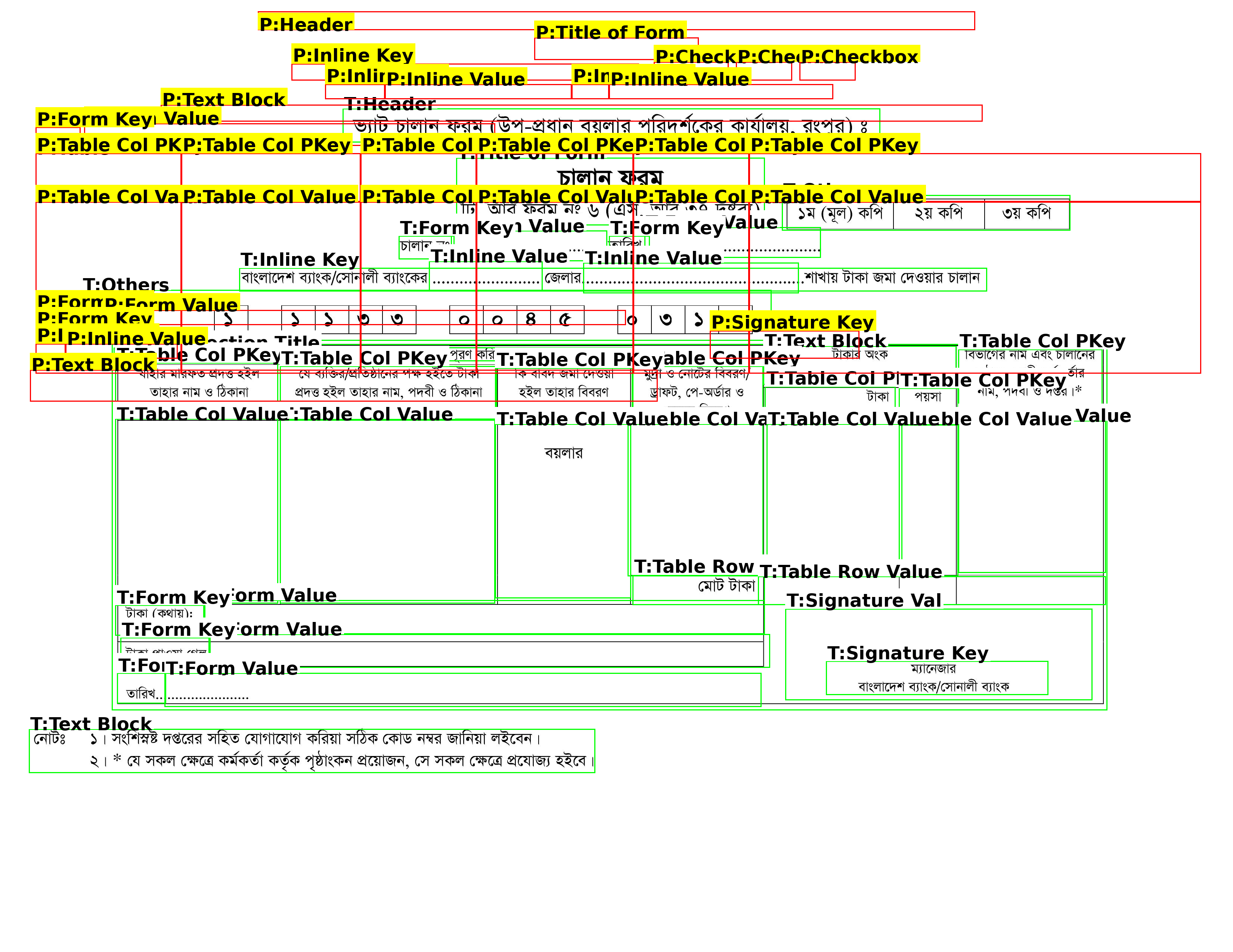}
        \caption{Shifted Predictions (\model{GPT-5.2})}
        \label{fig:gran_a}
    \end{subfigure}
    \hfill
    \begin{subfigure}{0.45\textwidth}
        \includegraphics[width=\linewidth]{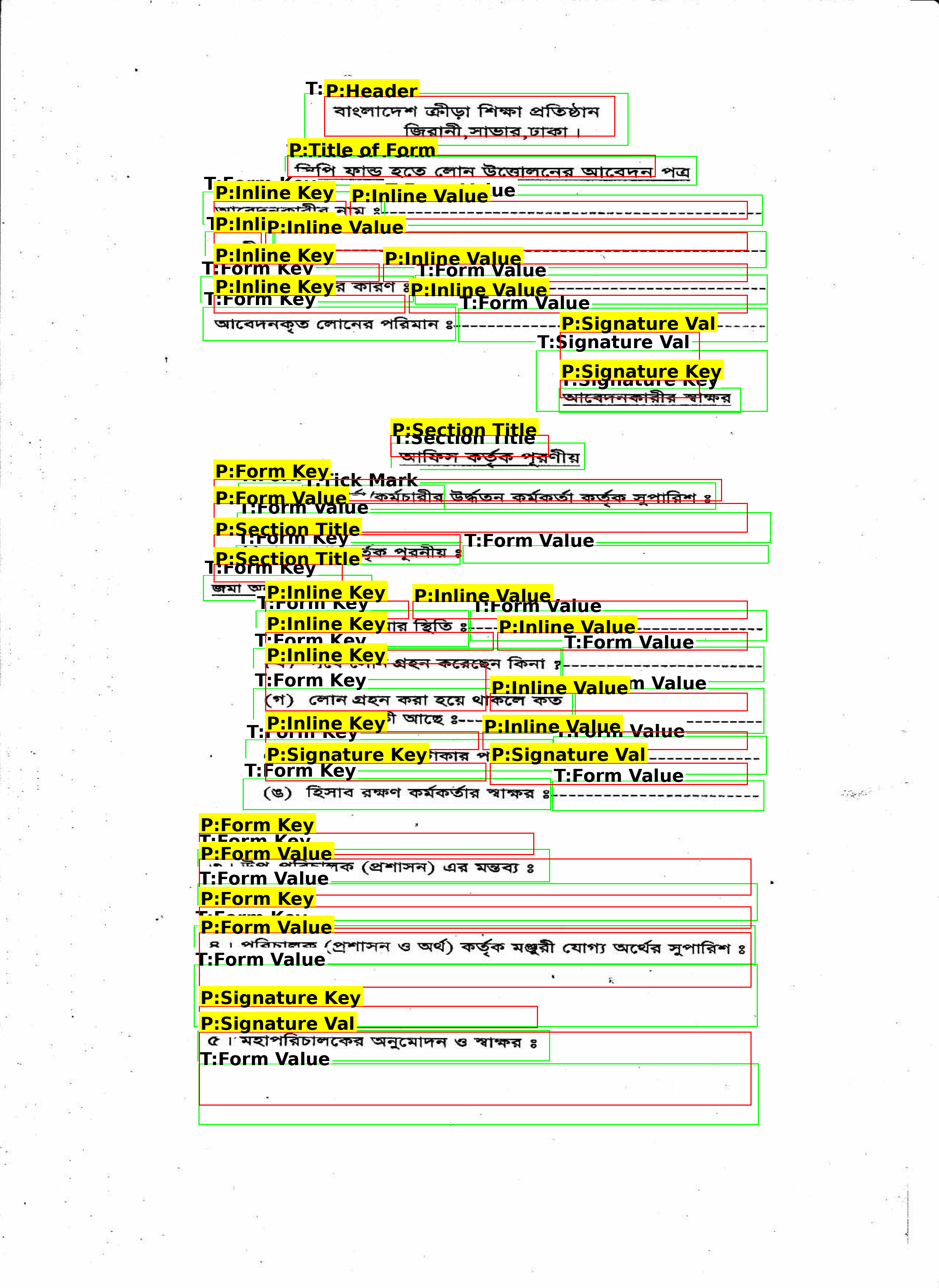}
        \caption{Miscategorized Form Entities (\model{Gemini 3 Pro})}
        \label{fig:gran_b}
    \end{subfigure}

    \begin{subfigure}{0.45\textwidth}
        \includegraphics[width=\linewidth]{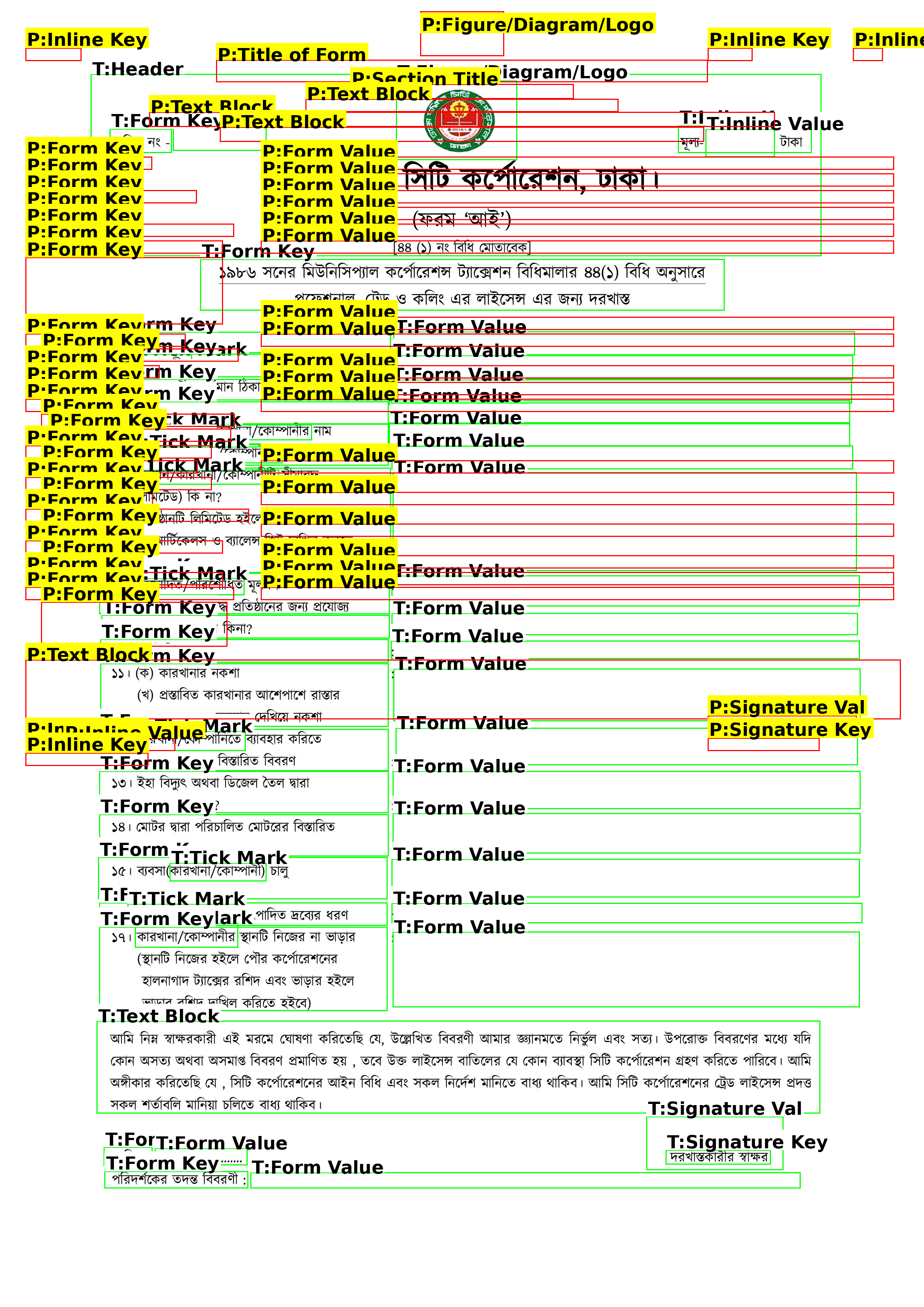}
        \caption{Hallucination (\model{Claude Opus 4.6})}
        \label{fig:gran_c}
    \end{subfigure}
    \hfill
    \begin{subfigure}{0.45\textwidth}
        \includegraphics[width=\linewidth]{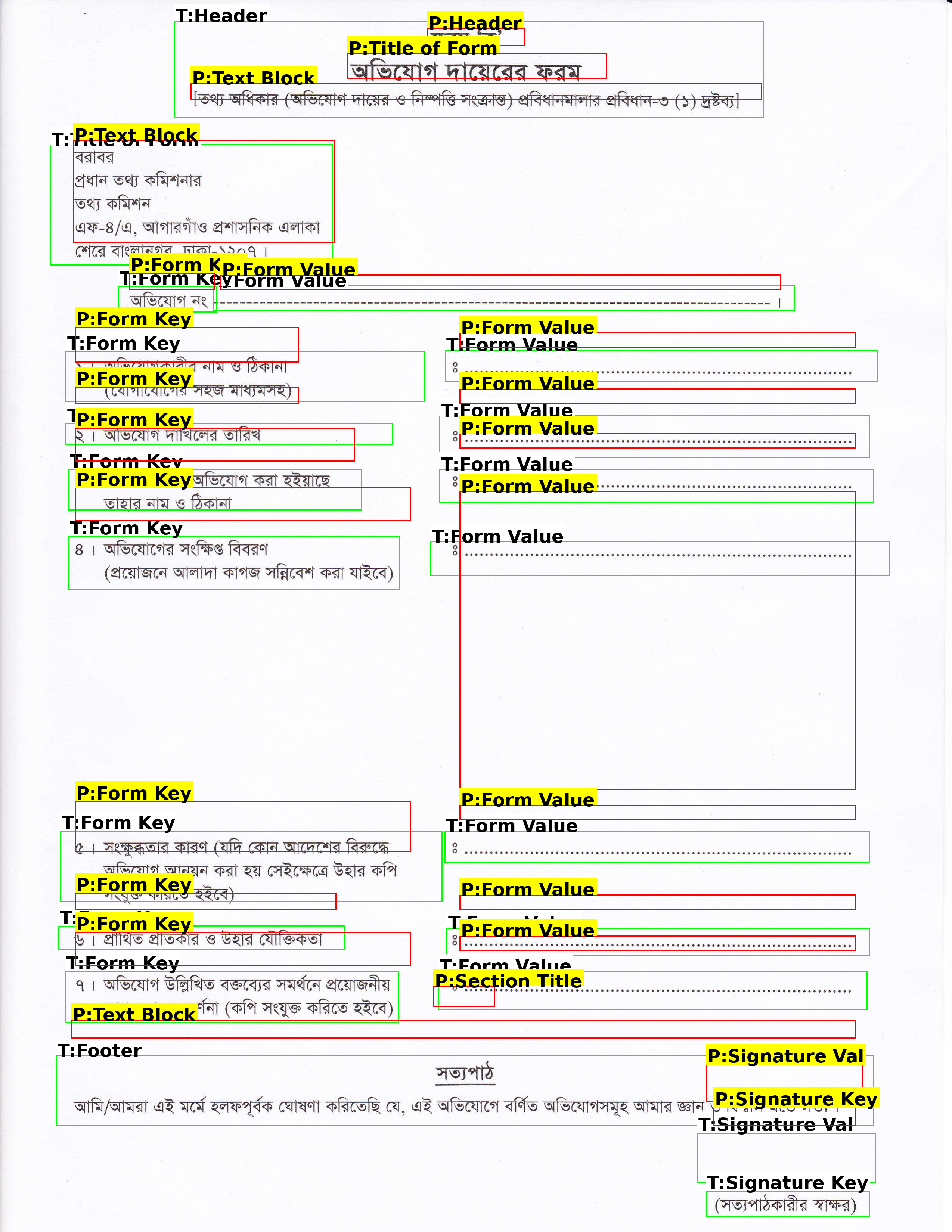}
        \caption{A Well Predicted Form (\model{Gemini 3 Pro})}
        \label{fig:gran_d}
    \end{subfigure}

    \caption{Failure and success modes of MLLM predictions for granular form entities.}
    \label{fig:all_granular}
\end{figure}
\begin{figure}[!htb]
    \centering
    \begin{subfigure}{0.45\textwidth}
        \includegraphics[width=\linewidth]{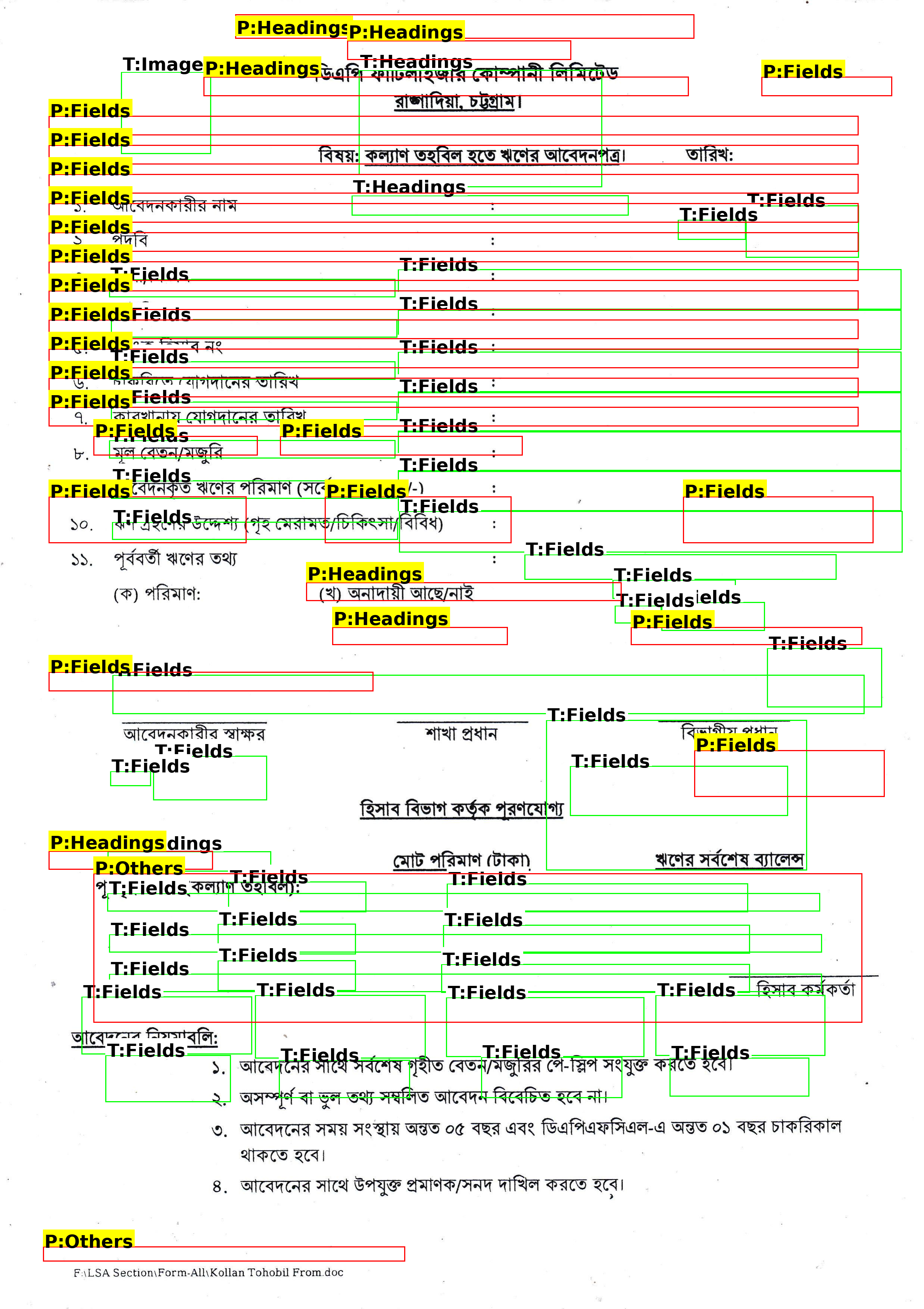}
        \caption{Partially Correct Predictions (\model{Claude Opus 4.6})}
        \label{fig:coarse_a}
    \end{subfigure}
    \hfill
    \begin{subfigure}{0.45\textwidth}
        \includegraphics[width=\linewidth]{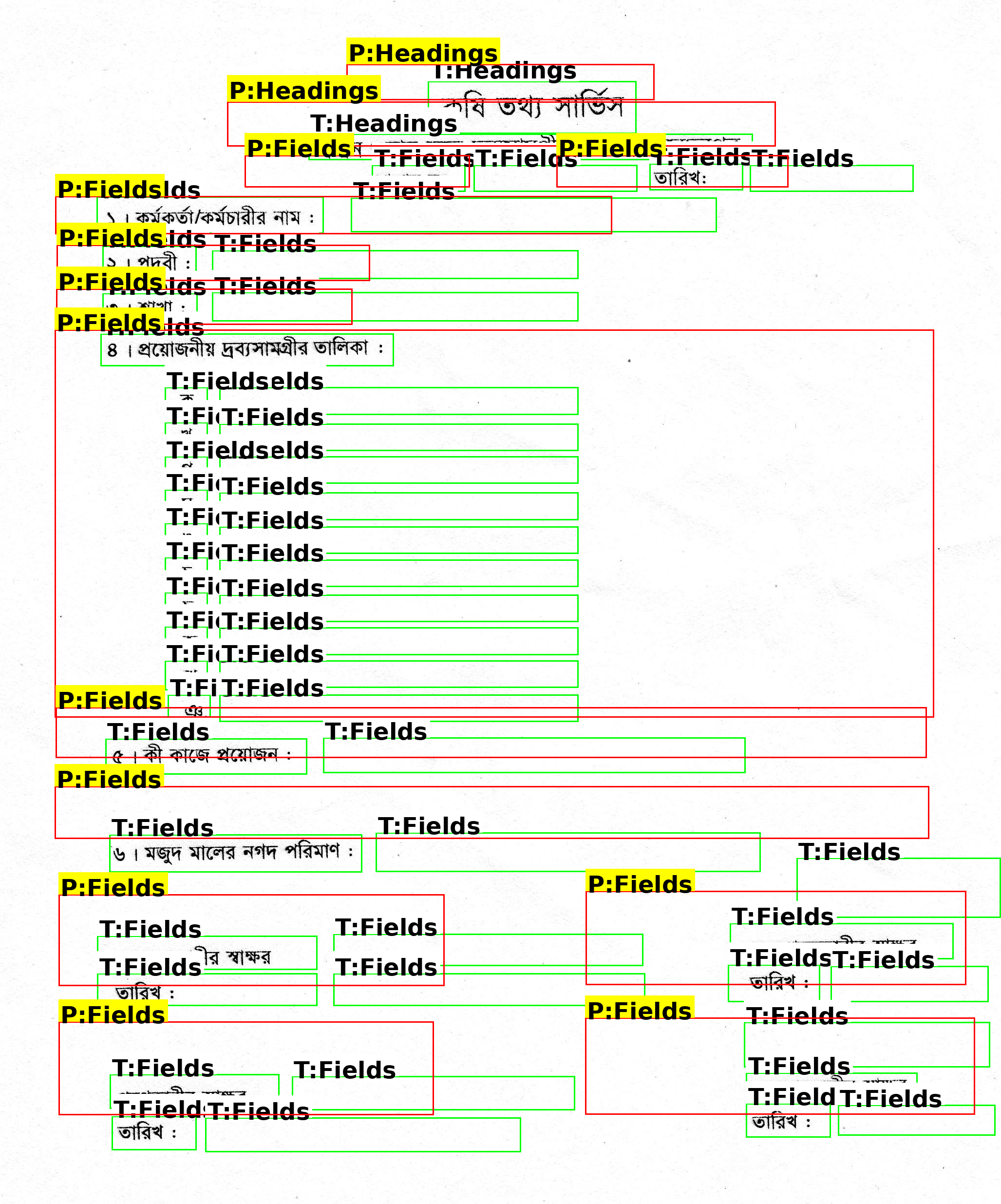}
        \caption{Sub-Field Prediction Failure (\model{Claude Opus 4.6})}
        \label{fig:coarse_b}
    \end{subfigure}
    \begin{subfigure}{0.45\textwidth}
        \includegraphics[width=\linewidth]{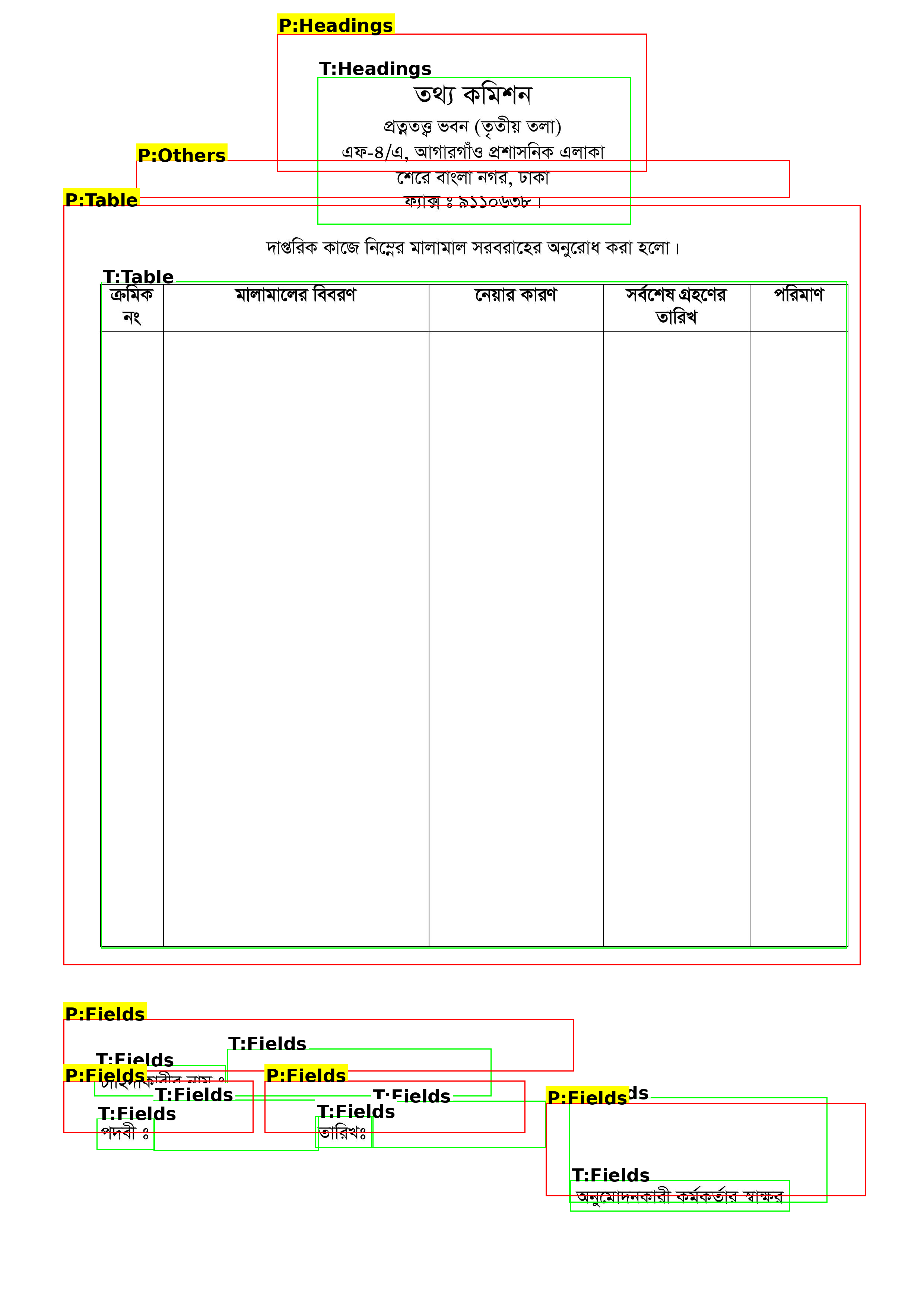}
        \caption{Prediction on a Easy Form (\model{GPT-5.2})}
        \label{fig:coarse_c}
    \end{subfigure}
    \hfill
    \begin{subfigure}{0.45\textwidth}
        \includegraphics[width=\linewidth]{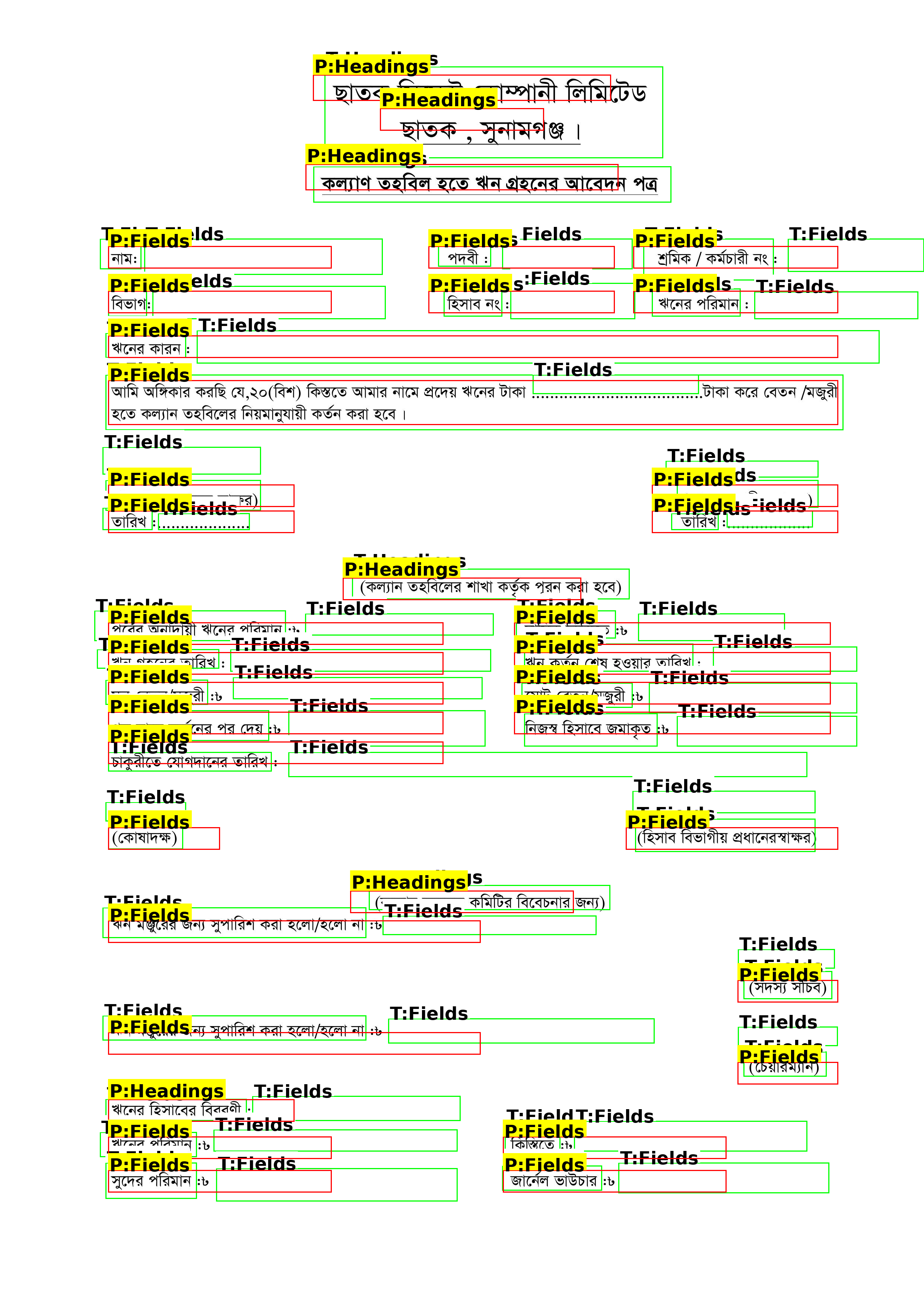}
        \caption{Prediction on a Complex Form (\model{Gemini 3 Pro})}
        \label{fig:coarse_d}
    \end{subfigure}
    \caption{Failure and success modes of MLLM predictions for coarse form entities.}
    \label{fig:all_coarse}
\end{figure}
\clearpage

\subsection{Key Information Extraction (KIE)}

\cref{fig:kie} presents representative success and failure cases for key information extraction on Bangla and English forms.

\textbf{Date extraction errors:} A recurring failure occurs when extracting dates from Bangla forms. As shown in \cref{fig:kie_a}, all evaluated MLLMs misread the date-of-birth field. The ground-truth value is \textit{15-06-2008}, whereas \model{Gemini 3 Pro} and \model{Claude} predict \textit{16-06-2008}, and \model{GPT-5.2} predicts \textit{18-06-2008}. Although structurally plausible, these predictions differ by one or more digits, suggesting that precise character-level recognition remains challenging when extracting Bangla date expressions.

\textbf{Numerical value hallucination:} Another failure mode arises when extracting numerical values. In \cref{fig:kie_c}, the ground-truth value is \textit{25000}, whereas \model{Gemini 3 Pro}, \model{GPT-5.2}, and \model{Claude} predict \textit{28000}, \textit{26000}, and \textit{24000}, respectively. None of these values appears anywhere in the document, indicating that the models hallucinate plausible numbers rather than grounding their predictions in the visual evidence.

\textbf{Successful extraction despite layout irregularities:} \cref{fig:kie_b} shows a representative success case. The \textit{Application for the period} field is visually irregular: the complete date range is written in the region corresponding to the \textit{From} field, while the \textit{To} field is empty. Despite this atypical layout, all evaluated MLLMs correctly extract the intended date range, demonstrating that they can exploit contextual cues even when the spatial organization deviates from the expected form structure.

\textbf{Minor spelling variations in Bangla text extraction:}
Another observed issue involves subtle character-level spelling errors in Bangla text fields. As shown in \cref{fig:kie_d}, \model{Gemini 3 Pro} replaces one Bangla consonant with another that has a similar pronunciation. Although the two characters are phonetically similar and often correspond to a similar \emph{sh} sound in English transliteration, the substitution results in an incorrect spelling of the extracted text. This suggests that MLLMs may rely on approximate phonetic representations when processing Bangla script, leading to minor but semantically meaningful transcription errors in extracted values.

\begin{figure}[htbp]
    \centering
    \begin{subfigure}{0.45\textwidth}
        \includegraphics[width=\linewidth]{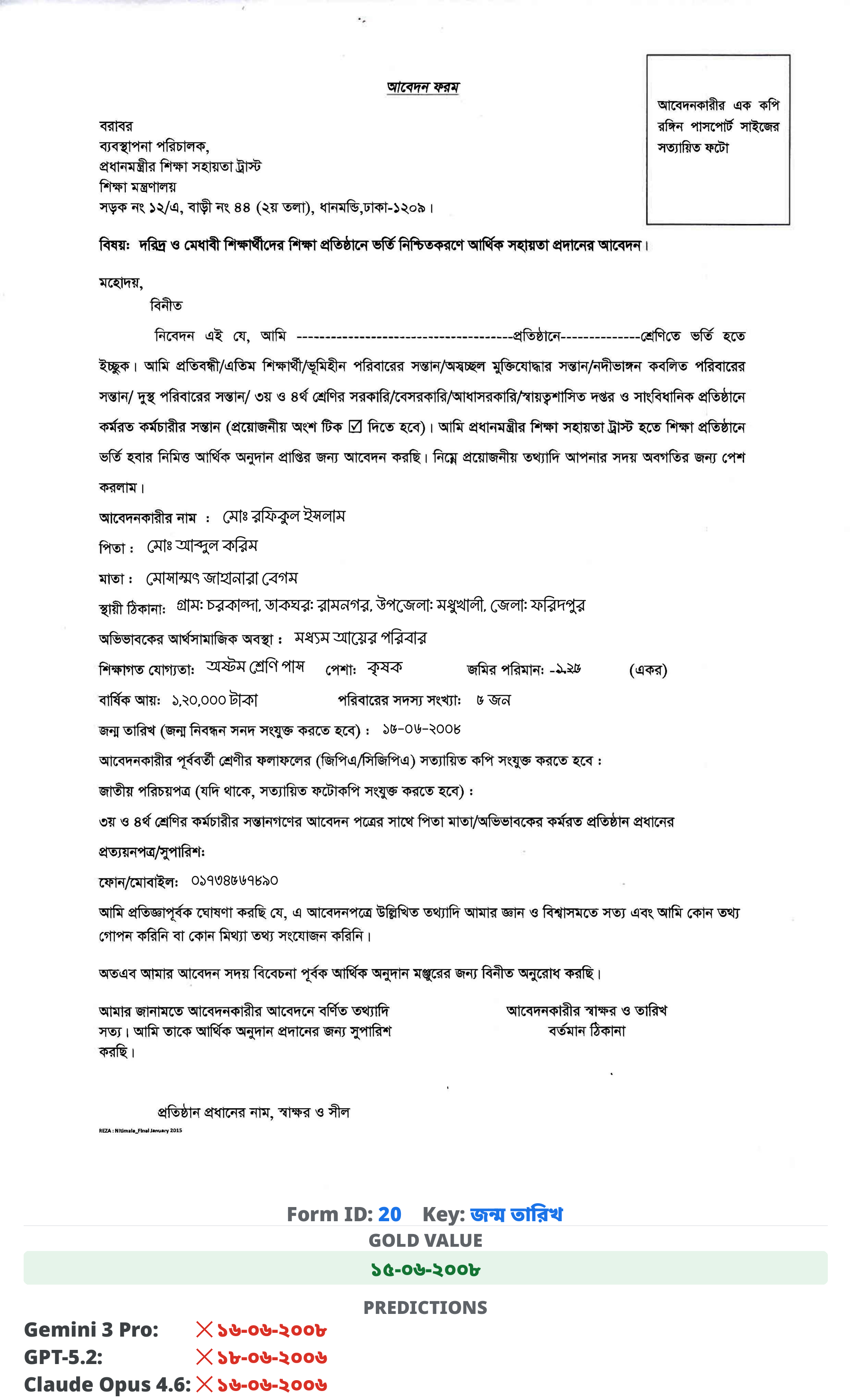}
        \caption{Wrong predictions of date for Bangla forms}
        \label{fig:kie_a}
    \end{subfigure}
    \hfill
    \begin{subfigure}{0.45\textwidth}
        \includegraphics[width=\linewidth]{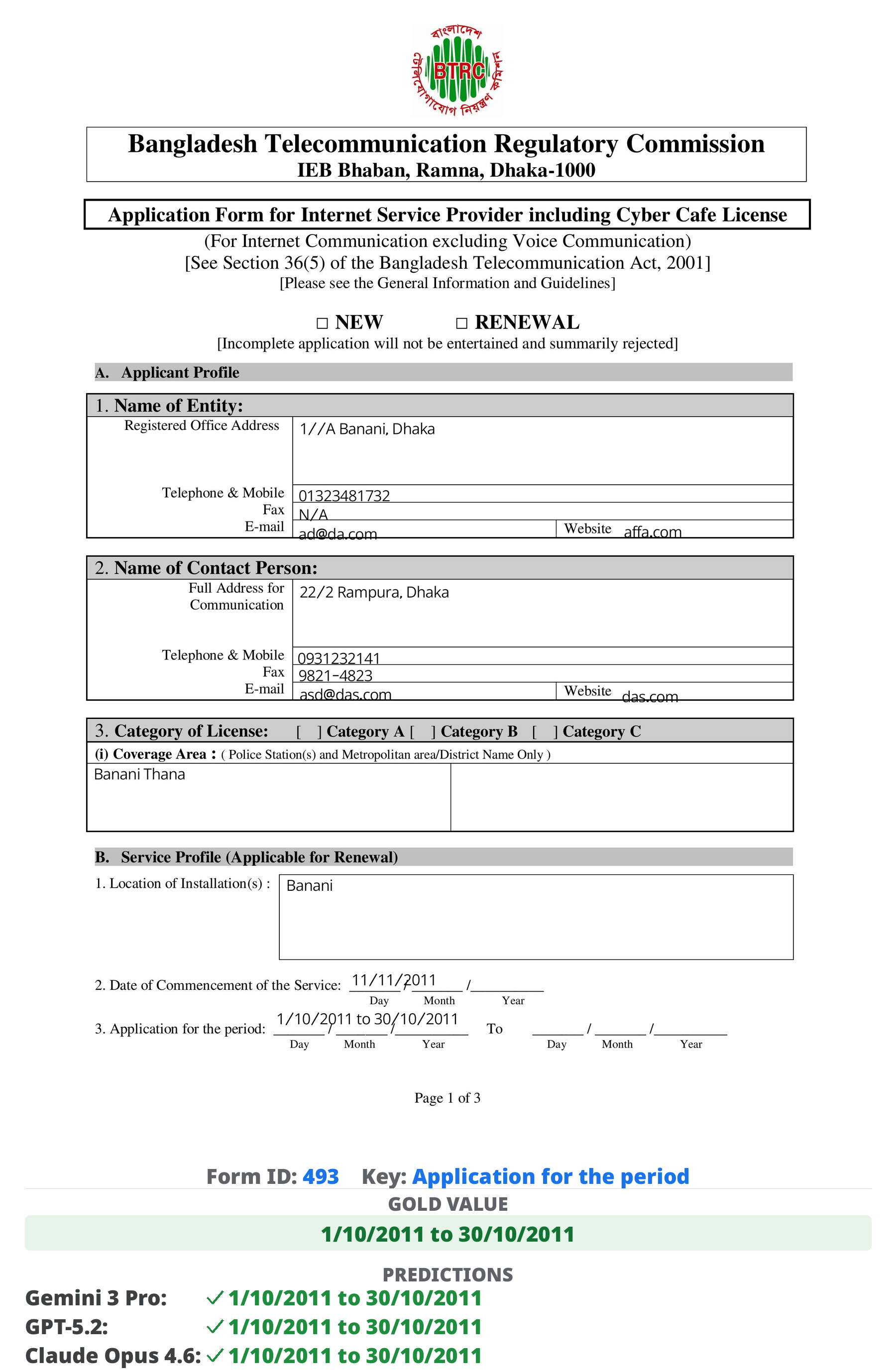}
        \caption{Correct predictions of date for English forms, even in inconsistent structure}
        \label{fig:kie_b}
    \end{subfigure}

    \begin{subfigure}{0.45\textwidth}
        \includegraphics[width=\linewidth]{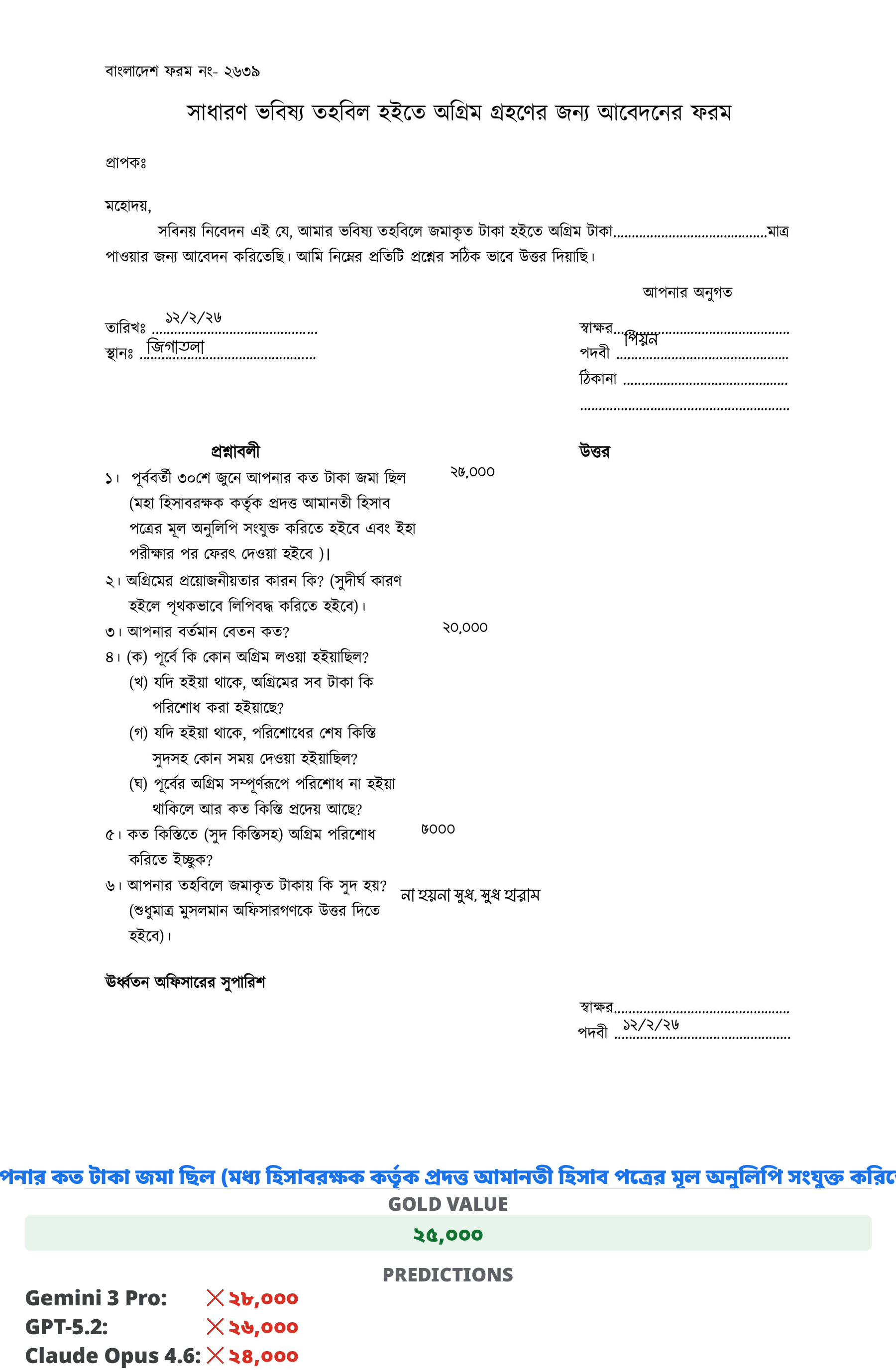}
        \caption{Numerical value hallucination in Bangla forms}
        \label{fig:kie_c}
    \end{subfigure}
    \hfill
    \begin{subfigure}{0.45\textwidth}
        \includegraphics[width=\linewidth]{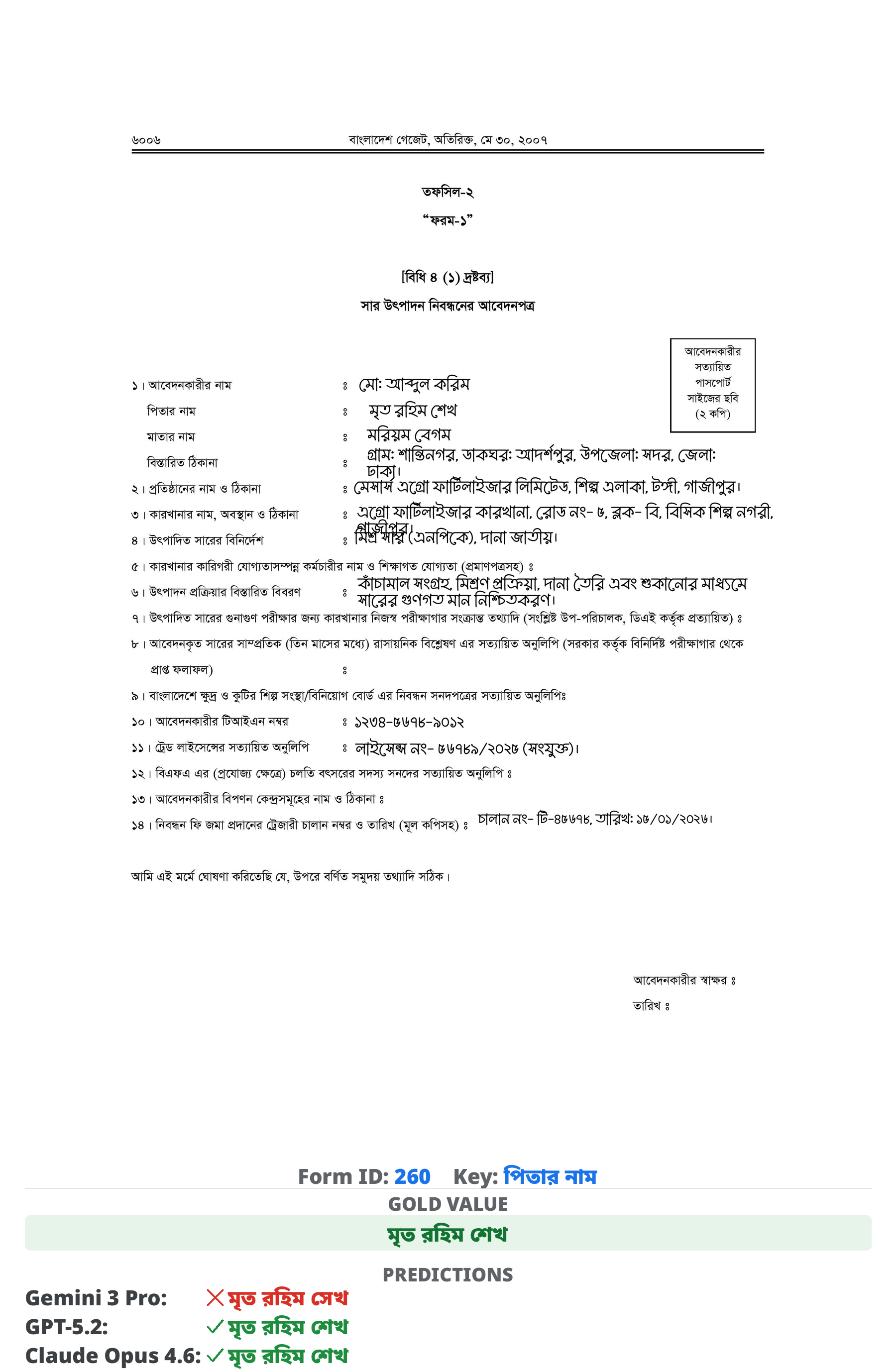}
        \caption{Minor spelling variations in Bangla text extraction}
        \label{fig:kie_d}
    \end{subfigure}

    \caption{Failure and success modes of MLLM predictions for KIE Task.}
    \label{fig:kie}
\end{figure}
\clearpage

\section{Prompts}
\label{sec:append_prompts}

\begin{equation}
    p_k = s_{\text{task}} \;\oplus\; s_{\text{inst}} \;\oplus\; s_k
\end{equation}

where
\begin{itemize}
    \item $s_{\text{task}}$ denotes the input, task, and output specification,
    \item $s_{\text{inst}}$ denotes the form-entity-specific instructions,
    \item $s_k$ denotes the prompt-specific instructions for prompt variant $k$.
\end{itemize}

The operator $\oplus$ denotes string concatenation. In the zero-shot setting, the prompt-specific component ($s_{k=\mathrm{zs}}$) is empty. In the chain-of-thought (CoT) setting, $s_{k=\mathrm{cot}}$ consists of the additional instruction:

\begin{quote}
\textit{Before producing the final output, internally reason step by step about the layout structure and label assignments, but do not include any intermediate reasoning in the output.}
\end{quote}

\subsection{Prompt Template Details}

In the prompt templates below, boldfaced placeholders (e.g., \textbf{\{form\_field\_categories\}}) denote variables, while all other braces are literal JSON syntax.

\begin{promptbox}{Core DLA Prompt ($s_{\mathrm{task}}$)}
  You are a **form layout understanding AI**. Given an image of a form page, your task is to analyze its layout and identify all visible fields along with their categories and bounding boxes.
  input:
  a. image of form page
  b. full image size (height and width)
  c. form field categories

  TASKS:
  1. Localize every visible field on the page.
  2. Classify each field into one of the given form field categories.
  3. Return bounding boxes [X, Y, W, H], where (X, Y) is the top-left corner, W is width, and H is height.

  OUTPUT FORMAT (exactly as shown):
  [
    {
      "id": "<unique_id>",
      "reasoning": <brief 1-line non-generic reasoning on choice of class>,
      "class": "<category_name>",
      "bbox": <Length 4 list [X, Y, W, H] where (X, Y) represents left-upper corner of bounding box and W and H width and height of the box respectively>,
      "confidence": <float 0-1>
    }
  ]

  COORDINATES:
  - Origin (0,0) = top-left; units = pixels.
  - Boxes tightly bound visible content, EXCEPT writable areas, which should include the full writable region.
  - Overlap only when nested.
  - Coordinates must be absolute pixel values expressed as floats (at most 4 decimal places).
  - Do NOT use repeating decimal patterns.

  |{entity_label_rules}|

  OVERALL QUALITY RULES:
  - Detect all visible fields; no omissions.
  - Use visual layout, proximity, and language.
  - Prefer meaningful, tightly bounded boxes.
  - Do not output a single object unless the page truly contains only one visible element.
  - Do not stop after the first detection if more than one visible element exists. Keep iterating until you have enumerated every visible field.
  - Use "Others" for uncertain cases.
  - Output must be in the described output format (a collection of JSON objects) - no other keys, natural text, or comments.

  Full Image Size: Height -> |{form_image_height}|, Width -> |{form_image_width}|
  Available Form Field Categories: |{form_field_categories}|
  Use only the category names exactly as provided in 'Available Form Field Categories'. Do not invent, abbreviate, or modify category names.
  Image: <image>
  |{cot_instruction}|
  Your Layout Analysis Output:
  \end{promptbox}
\begin{promptbox}{Coarse Entity Rules ($s_{\mathrm{inst}}$, 5 classes)}
  FIELD RULES:
  - **Headings**: page-level structural text such as headers, footers, titles, or section headers.
  - **Fields**: any region related to user input, including both labels requesting information (keys) and the corresponding writable areas (values), as well as inline key-value blanks, checkboxes, tick
  marks, signature labels, signing areas, photo attachment placeholders, and other fillable elements.
  - **Image**: purely graphical elements such as logos, icons, figures, or diagrams that are not intended for user input. Photo attachment placeholders are fields, not Image.
  - **Table**: tabular structures with rows and/or columns, even if grid lines are faint or implicit.
  - **Others**: instructional text, page numbers, scribbles, stray marks, or uncertain regions.

  TABLE Specific RULES:
  A. STRUCTURE
  - Draw exactly one **Table** bounding box around the full table/grid region.
  - Do NOT output internal rows, columns, headers, or cells.

  B. RELATIONSHIPS
  - Table captions or identifiers that are visually integrated with the table may be included inside the Table bounding box.

  C. INVISIBLE GRIDS
  - Infer alignment from consistent gutters or repeated leaders.
  - Keep table edges straight and globally aligned.
  \end{promptbox}
\begin{promptbox}{Granular Entity Rules ($s_{\mathrm{inst}}$, 26 classes)}
  FIELD RULES:
  - **Header, Footer, Title of Form, Section Title, Page Num**: structural bands or labels.
  - **Form Key**: label requesting input ("Name", "Date of Birth").
  - **Form Value**: blank region for user entry; spacious bounding box.
  - **Inline Key, Inline Value**: key phrases and blanks within a sentence; value = writable area.
  - **Checkbox, Tick Mark**: selectable elements ("$\square$ Male", "Yes / No").
  - **Signature Key, Signature Val**: signature label and signing area (spacious box).
  - **Photo Field**: placeholders for attaching photos
  - **Figure/Diagram/Logo**: icons. (This is one category covering figures, diagrams, and logos)
  - **Text Block**: instructions or narrative text.
  - **Gibberish, Mark for removal**: scribbles, stray marks.
  - **Others**: fallback for unclassified regions.

  TABLE RULES:
  A. STRUCTURE
  - Draw one **Table** box around the full grid region (even if lines are faint).
  - Choose one decomposition style:
  - **Column-major** $\rightarrow$ columns have headers at the top.
  - **Row-major** $\rightarrow$ rows have labels on the left.
  Do not mix both for one table.

  B. RELATIONSHIPS
  - Internal elements (keys, values, captions, section titles) set parent_id to the Table.
  - **Table Caption**: title above/below the table.
  - **Table Index**: short identifier referencing the table (e.g., "Table 1", "Table II", "Exhibit A"), usually placed immediately before the caption.
  - **Table Section Title**: header band dividing row/column groups.

  C. COLUMN-MAJOR
  - **Table Col PKey**: header cell areas across the top.
  - **Table Col Value**: vertical strips below headers spanning to table bottom.
  - Merged headers $\rightarrow$ one PKey linked to multiple Value regions
  - No overlap among adjacent columns.

  D. ROW-MAJOR
  - **Table Row PKey**: leftmost identifying cells for each row.
  - **Table Row Value**: remaining cells of that row.
  - Row Value spans full row height for multi-line content.

  E. MERGED / IRREGULAR CELLS
  - Assign each merged cell as key or value based on function.
  - Adjust boxes to include full merged area.

  F. INVISIBLE GRIDS
  - Infer alignment from consistent gutters or repeated leaders.
  - Keep column/row edges straight and globally aligned.

  G. CONSISTENCY
  - Keys = label regions; Values = data regions.
  - No overlaps among peers.
  - Preserve inner whitespace for writable cells.

  H. SECTION TITLES INSIDE TABLES
  - Bold or shaded header rows spanning multiple columns $\rightarrow$ Table Section Title.

  I. MULTI-PAGE TABLES
  - Annotate each page separately; repeat headers when they reappear.

  J. TEXT PREVIEW
  - Include short snippet if legible ("Name", "Total"), else empty string.
  \end{promptbox}
\begin{promptbox}{Prompt-Specific Instruction ($s_{k=\mathrm{cot}}$)}
Before producing the final output, internally reason step by step about the layout structure and label assignments, but do not include any intermediate reasoning in the output.
\end{promptbox}
\begin{promptbox}{Zero-Shot KIE Prompt}
  You are an information extraction system.
  Given:
  1. An image of a document.
  2. A question in the format: What is the value of <key_text>?

  Your task:
  - Locate the field corresponding to <key_text>.
  - Extract the exact visible value from the document.
  - If multiple possible values are found, return the one closest to the key text in the document layout.

  Output rules:
  - Return the answer strictly inside these tags:
    <answer>the exact extracted value</answer>
  - If the key is not present:
    <answer>NOT_FOUND</answer>
  - If the value is blank:
    <answer>EMPTY</answer>
  - Do NOT output anything outside these tags.
  - Do NOT explain.
  - Do NOT add extra text.

  Important:
  - Extract exactly as written (preserve formatting, symbols, punctuation).
  - Do NOT translate the value.
  - Do NOT switch the language of the extracted text.
  - Return the value exactly in the original script and language as it appears in the document.
  - Do not infer missing information.
  - Do not calculate or transform values.

  Now answer the following question using the rules above.

  Question: |{question}|
  Your answer:
  \end{promptbox}

%\subsubsection{DLA Zero-Shot Evaluation Prompt Template for Full Entity Set}
%\label{sec:zsprompt-template}
%\input{res/prompts/zs-prompt}

%\subsubsection{DLA Zero-Shot Evaluation Prompt Template for Reduced Entity Set}
%\label{sec:zsprompt-template_reduced}
%\input{res/prompts/reduced_dla_zs}

%\subsubsection{DLA Chain-of-Thought (CoT) Evaluation Prompt Template for Full Entity Set}
%\label{sec:cotprompt-template}
%\input{res/prompts/cot-prompt}

%\subsubsection{DLA Chain-of-Thought (CoT) Evaluation Prompt Template for Reduced Entity Set}
%\label{sec:cotprompt-template}
%\input{res/prompts/reduced_dla_cot}

%\subsubsection{KIE Evaluation Prompt Template}
%\label{sec:kie_zs}
%\input{res/prompts/kie_zs_prompt} 

\end{document}